\documentclass{article}
\usepackage{iclr2027_conference,times}

\usepackage[utf8]{inputenc}
\usepackage[T1]{fontenc}
\usepackage[hidelinks,hypertexnames=false]{hyperref}
\usepackage{url}
\makeatletter
\g@addto@macro\UrlBreaks{\UrlOrds}
\makeatother
\usepackage{booktabs}
\usepackage{etoolbox}
\AtBeginEnvironment{table}{\setlength{\belowcaptionskip}{6pt}}
\usepackage{amsmath,amsfonts,amssymb}
\usepackage{microtype}
\usepackage{xcolor}
\usepackage{graphicx}
\usepackage{array}
\usepackage{enumitem}
\usepackage{float}
\graphicspath{{./}{figures/}}

\newcolumntype{L}[1]{>{\raggedright\arraybackslash}p{#1}}
\newcommand{\cai}{\operatorname{CAI}}
\newcommand{\damage}{\operatorname{D}}
\newcommand{\acc}{\operatorname{acc}}
\newcommand{\ws}{\mathrm{ws}}
\newcommand{\content}{\mathrm{cont}}
\newcommand{\mb}{\mathrm{mb}}
\newcommand{\rs}{\mathrm{rs}}
\newcommand{\Rdom}{\mathcal{R}}
\newcommand{\Ndom}{\mathcal{N}}

\newcommand{\DeltaChars}{\Delta_{\mathrm{chars}}}
\newcommand{\DeltaWS}{\Delta_{\mathrm{ws}}}
\newcommand{\secref}[1]{Section~\ref{#1}}
\newcommand{\tabref}[1]{Table~\ref{#1}}
\newcommand{\figref}[1]{Figure~\ref{#1}}

\title{Target-Token Asymmetry Can Masquerade as Depth Specialization}
\iclrfinalcopy
\makeatletter
\patchcmd{\@maketitle}
  {Published as a conference paper at ICLR 2027}
  {Preprint}
  {}{\PackageError{arxiv-preprint}{Could not set the preprint header}{Check the local style file.}}
\makeatother
\newcommand{\PaperAuthors}{Danlin Li, Miao Li, Xu Han, Jiecheng Lu}
\author{%
Danlin Li$^{1}$\quad Miao Li$^{1}$\quad Xu Han$^{2}$\quad Jiecheng Lu$^{1}$\\
\normalfont $^{1}$Georgia Institute of Technology\quad $^{2}$AWS\\
\normalfont\texttt{jlu414@gatech.edu}%
}

\hypersetup{
  pdftitle={Target-Token Asymmetry Can Masquerade as Depth Specialization},
  pdfauthor={\PaperAuthors},
  pdfsubject={Language-model depth attribution and target-token composition},
  pdfkeywords={language models, interpretability, ablation, tokenization}
}

\begin{document}
\maketitle

\begin{abstract}
Layer-ablation studies often perturb early or late layers of a language model,
compare the damage to prediction on two corpora, and read a larger early-layer
effect on one corpus as early-layer specialization for its task. Such a
comparison averages over the tokens that each corpus asks the model to predict,
and these target tokens differ: math solutions, for example, ask for far more
whitespace and numbers than encyclopedia prose. We show that this difference can
create the appearance of depth specialization. We split a cross-corpus
early--late damage interaction by target-token type and recompute it under a
target mixture shared by both corpora. Between GSM8K and Wikitext, a positive interaction replicates across
five model instances. In the three fully decomposed models, only
whitespace has a positive interaction. Removing whitespace reverses
the pooled effect in all three, and matching the full target mixture
reverses it in two. The diagnosis holds under position matching, adversarial reassignment of
token types, and a continuous readout. The risk can also be anticipated from the
tokenizer alone. A screen computed before any model runs places every held-out
pair it scored in the region its label predicts, and on the pairs it rates at
high or medium risk, removing whitespace reverses every positive pooled estimate.
A smaller content-word effect one token ahead follows the share of content
targets that the unablated model already predicts. A cross-corpus depth effect
therefore supports an interpretation of task specialization only after the target-token mixture is
reported, the effect is decomposed by token type, and a mixture-matched summary
keeps its sign.
\end{abstract}

\section{Introduction}
\label{sec:intro}

Interpretability research often asks which layers of a Transformer
\citep{vaswani2017attention} carry the computation behind a behavior, because the
answer shapes mechanistic explanations and tells later analyses where to look. A
common and inexpensive test perturbs a band of early or late layers, measures how
much next-token accuracy drops, and compares the drop across corpora. Suppose
that perturbing the early layers hurts accuracy on math solutions more than on
encyclopedia prose, relative to the same comparison for the late layers. It is
then natural to conclude that the early layers matter more for mathematical
reasoning, and the conclusion gains weight when the pattern repeats across model
families.

\begin{figure}[t]
\centering
\includegraphics[width=\linewidth]{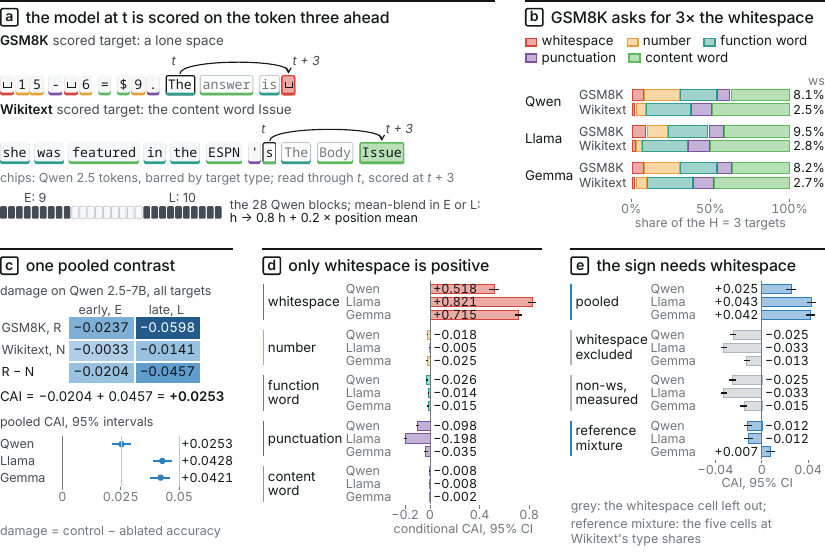}
\caption{The argument on one model instance: Qwen2.5-7B, GSM8K against Wikitext,
three tokens ahead, mean-blend, top-1 accuracy. Panel \textbf{a} quotes two real
evaluated positions, one chip per token with the scored target filled, selected
for readability and target type (Appendix~\ref{app:vignettes}), and marks the two
blended layer bands. Panel \textbf{b} gives each corpus's H=3 target-type shares
in all three models. Panel \textbf{c} fills in the four damage terms of
Eq.~\eqref{eq:cai} and gives each model's pooled CAI. Panel \textbf{d} gives the
conditional CAI of each target type, Eq.~\eqref{eq:condcai}, and panel \textbf{e}
sets the pooled estimate beside two summaries $M(q)$ and the directly measured
non-whitespace CAI. Under the two corpora's combined mixture $q$, with whitespace
share $5.3\%$, $M(q)=0.053\times0.518-0.023=+0.004$. The pooled $+0.0253$ weights
each corpus by its own mixture: $M(\pi_\Rdom)=+0.0207$, the mixture shift adds
$+0.0057$, and per-sequence averaging adds $-0.0011$. Intervals are 95\%,
and \tabref{tab:canonical} lists the values.}
\label{fig:idea}
\end{figure}

This reading assumes that the two corpora pose comparable prediction tasks. The
token that the model must predict at an evaluated position is its target token,
and a pooled accuracy averages over every target, whether it is a word, a number,
a punctuation mark, or a bare space. Corpora differ in how often each kind occurs.
\figref{fig:idea}a shows two evaluated positions. In a GSM8K solution
\citep{cobbe2021gsm8k} that ends with ``The answer is 9'', the tokenizer writes
the space before the number as a token of its own, so the model is scored on
predicting a lone space. At an evaluated position in a Wikitext article
\citep{merity2016wikitext}, the model is scored on the word ``Issue''. Math
solutions ask for far more whitespace and numbers than encyclopedia prose does
(\figref{fig:idea}b). A perturbation that changes how often the model predicts a
space therefore moves the pooled score of the math corpus more, even if it
affects words in the same way in both corpora.

This confound is hard to see because it lives inside the quantity that makes the
test attractive. A pooled score is cheap to compute and easy to replicate, and a
target mixture that repeats across models makes the pooled contrast repeat as
well, so replication alone cannot tell a depth effect from a composition effect.
Aggregation bias of this kind is classical \citep{simpson1951}. Work on
tokenization disparity \citep{petrov2023tokenizers,wegmann2025tokenization},
evaluation validity \citep{liang2023helm,schaeffer2023mirage}, and the dependence
of activation patching and ablation on the chosen intervention and metric
\citep{zhang2024activationpatching,li2024optimalablation,miller2024faithfulness}
documents related measurement risks. The stakes are practical: layer-level
findings guide where later work edits a model \citep{meng2022rome} or which layers
it decodes from. For example, DoLa contrasts predictions from earlier and later layers
to improve factuality \citep{chuang2024dola}. Cross-corpus depth attribution therefore needs a direct measurement of how much of a pooled contrast the target mixture carries, and a way to anticipate that risk before a model is run.

We study one such contrast in detail. Our measure, the contextual ablation
interaction (CAI), is a difference of differences: how much more an early-layer
perturbation damages accuracy on a focal corpus than on a reference corpus, minus
the same difference for a late-layer perturbation (\figref{fig:idea}c). We fix a
mild perturbation that blends each layer's output toward the corpus average at
the same position, and we score the model on the token three positions past the visible
text. In this setting the CAI of GSM8K against Wikitext is positive in every model
instance we audited. We then split the CAI by the type of the target token,
recompute it under a target mixture shared by both corpora, and test whether a
screen computed from the tokenizer alone can label a corpus pair's risk before any
model runs.

The positive contrast comes from whitespace targets (\figref{fig:idea}d--e). The
CAI restricted to whitespace targets is far larger than the pooled estimate,
every other target type has a non-positive estimate, and removing whitespace
reverses the pooled sign. Across the other high- and medium-risk corpus pairs, excluding
whitespace reverses every positive pooled estimate. Position
matching, adversarial reassignment of token types, and a continuous
readout further support the whitespace finding. On every held-out and DoLa pair that the screen rated at high or low
risk in advance, the measured interaction falls in the region that the label
predicts.

The paper makes three contributions. The first is a diagnosis: the evaluated
target population carries a replicated pooled depth contrast, which makes
target-token composition a concrete source of aggregation bias in cross-corpus
depth attribution. The second is a tokenizer-only screen that rates a corpus
pair's risk before any model runs. The third is a reporting protocol that
decomposes a pooled depth effect by target type, tests mixture-matched summaries,
and splits a remaining positive cell by whether the unablated model predicts its
targets. Applied to a content-word effect one token ahead under a second
perturbation, that split traces the effect to how many content targets the model
already predicts in each corpus.

\section{Measuring a depth contrast by target type}
\label{sec:method}

\paragraph{Damage and the interaction.}
We compare a focal corpus $\Rdom$ with a reference corpus $\Ndom$ and perturb
either an early layer band $E$ or a late layer band $L$. At each evaluated
position $t$, the model reads the prefix $x_{1},\dots,x_{t}$, and the argmax of
its next-token logits at $t$ is scored against the target token $x_{t+k}$. Causal
masking hides $x_{t+1},\dots,x_{t+k}$, so the horizon H=$k$ asks for a token $k$
positions past the visible context, and H=1 is ordinary next-token prediction.
For a corpus $C$, operator $o$, and layer band $B$, intervention damage is
\begin{equation}
\damage^o_B(C,k)=\acc_{\rm ctrl}(C,k)-\acc^o_{\rm ablate}(B,C,k).
\end{equation}
Positive damage means that the intervention reduced accuracy relative to the
unperturbed control pass. The contextual ablation interaction compares the
excess damage on the focal corpus under the two bands:
\begin{equation}
\label{eq:cai}
\cai^o_k=[\damage^o_E(\Rdom,k)-\damage^o_E(\Ndom,k)]
          -[\damage^o_L(\Rdom,k)-\damage^o_L(\Ndom,k)].
\end{equation}
A positive CAI means that the early band creates more excess damage on $\Rdom$
than the late band does, the pattern that would suggest early-layer
specialization for the focal corpus's task. We split layers into thirds and use
the first and last thirds as $E$ and $L$. Our canonical pair is GSM8K as $\Rdom$
against Wikitext as $\Ndom$, and \figref{fig:idea}c fills in its four damage terms.

\paragraph{Interventions and readouts.}
Both interventions blend the residual-stream output of every Transformer block
in the band, at every position, as $h'=(1-\alpha)h+\alpha g$ with $\alpha=0.2$.
Mean-blend sets $g$ to $\bar h_{\rm pos}$, the same layer's control-pass mean at
the same position over the corpus's evaluated sequences. Mild-resample sets $g$
to the same layer and position of a donor sequence from the same corpus. Donors
follow a random derangement, so no sequence is its own donor, and positions
past the donor's length are left unchanged. The primary analysis uses H=3
mean-blend top-1 CAI, and the content-word cell of \secref{sec:surviving} uses
H=1 mild-resample. We also use the target margin, the gold-token logit minus the
largest competing logit, as a continuous readout.

\paragraph{Target types and conditional cells.}
A deterministic map classifies each vocabulary token by ordered rules on its
decoded string with surrounding whitespace stripped. The token is whitespace if
nothing remains; a number if every remaining character is a digit, period, comma,
or plus sign and at least one is a digit; punctuation if every character is
Unicode punctuation, a space, or a tab; a function word if the lower-cased string
is one of 106 listed function words; and a content word otherwise, including a
token that fails to decode. A leading space therefore
leaves a word's class unchanged, and a token mixing letters and digits is a
content word. Conditional CAI
replaces each accuracy in \eqref{eq:cai} with the mean per-sequence accuracy on
the targets of one type $\tau$, giving four conditional damage terms
$\damage^o_B(C,k\mid\tau)$:
\begin{equation}
\label{eq:condcai}
\cai^o_{k,\tau}=[\damage^o_E(\Rdom,k\mid\tau)-\damage^o_E(\Ndom,k\mid\tau)]
                -[\damage^o_L(\Rdom,k\mid\tau)-\damage^o_L(\Ndom,k\mid\tau)].
\end{equation}
A sequence enters the cell of type $\tau$ when its targets include that type,
and a cell requires at least 10 such sequences per corpus.

\paragraph{Mixture summaries.}
Let $\pi_C(\tau)$ be the share of type $\tau$ among corpus $C$'s evaluated
targets. For any mixture $q$ applied to both corpora, we summarize the cells as
$M(q)=\sum_\tau q(\tau)\cai^o_{k,\tau}$. \emph{Whitespace excluded} sets $q(\ws)=0$ and includes only the
other token types that meet the minimum sequence requirement.
We rescale their weights to sum to one. \emph{Reference mixture}
uses the reference corpus's target-type proportions, $q=\pi_\Ndom$.
For the canonical pair, the whitespace-excluded weights are based
on the two corpora's combined target counts. For the other corpus
pairs in \secref{sec:coverage}, they are based on the focal corpus's
counts from the tokenizer-only screen
(Appendix~\ref{app:decomposition}).
Pooled CAI reflects each corpus's own target-token mixture, whereas
$M(\pi_\Rdom)$ uses the focal corpus's mixture for both corpora.
Let $\Delta_C(\tau)=\damage^o_E(C,k\mid\tau)-\damage^o_L(C,k\mid\tau)$
denote the early--late difference in damage for type $\tau$.
If both pooled and conditional accuracies are calculated over tokens,
then
$\cai^o_k=M(\pi_\Rdom)+\sum_\tau[\pi_\Rdom(\tau)-\pi_\Ndom(\tau)]\Delta_\Ndom(\tau)$.
Our estimator instead averages accuracy within each sequence first,
which introduces a residual reported in Appendix~\ref{app:protocol}. All intervals are 95\% percentile intervals
from resampling sequences within each corpus. For a mixture summary every draw
recomputes the cells. The canonical draws also recompute the weights, and the
draws for the other pairs keep the screen's weights fixed.

\paragraph{Diagnostic interaction.}
To compare corpus pairs by how strongly a setting singles out whitespace, we
contrast the whitespace-minus-content gap of the primary setting with the same
gap in a reference setting, H=1 mild-resample, whose whitespace and content cells
are both small in every audited instance (\tabref{tab:app_audit_cells}). Writing
$\mb$ and $\rs$ for conditional CAI under mean-blend and mild-resample, the
diagnostic interaction is
\begin{equation}
\label{eq:audit}
I=(\mb_{H=3,\ws}-\mb_{H=3,\content})
 -(\rs_{H=1,\ws}-\rs_{H=1,\content}).
\end{equation}
Both terms come from model outputs, and a large positive $I$ marks a pair whose
primary-setting contrast is concentrated on whitespace.

\paragraph{Screen, data, and models.}
Before model evaluation, we measure two tokenizer-only quantities: the target
whitespace-rate gap $\DeltaWS=\pi_\Rdom(\ws)-\pi_\Ndom(\ws)$ and the
characters-per-token gap $\DeltaChars$, the reference corpus's mean characters
per token minus the focal corpus's. \secref{sec:coverage} states the rule that
turns them into a risk label before any model runs. The evaluated corpora include GSM8K
\citep{cobbe2021gsm8k}, Wikitext \citep{merity2016wikitext}, MATH
\citep{hendrycks2021math}, DROP \citep{dua2019drop}, ARC-Challenge
\citep{clark2018arc}, StrategyQA \citep{geva2021strategyqa}, FACTOR-News
\citep{muhlgay2024factor}, HotpotQA \citep{yang2018hotpotqa}, AG News
\citep{zhang2015agnews}, SQuAD \citep{rajpurkar2016squad}, TinyStories
\citep{eldan2023tinystories}, and CodeSearchNet \citep{husain2019codesearchnet}.
The three primary model families are Qwen2.5, Llama 3.2, and Gemma 2
\citep{qwen2024qwen25,grattafiori2024llama3,gemmateam2024gemma2}. The canonical
five-instance audit also includes gemma-2-9b-it and Hermes-3-Llama-3.1-8B
\citep{teknium2024hermes3}. Except in \tabref{tab:app_base}, which lists base checkpoints, every model name denotes an instruction-tuned checkpoint. Appendix~\ref{app:protocol} lists the short names used in figure panels. \tabref{tab:coverage} gives per-pair sample sizes.

\section{The pooled effect collapses to whitespace}
\label{sec:result1}

We start with the canonical pair in the primary setting, H=3 mean-blend top-1
accuracy, for the three primary models (\tabref{tab:canonical}). The pooled CAI
is small and positive, while the whitespace cell is larger than it by more than
an order of magnitude in every model. The four non-whitespace point estimates are
non-positive in every model. Eleven of those twelve intervals lie strictly below
zero, and the Gemma content interval includes zero, with an upper endpoint of
$+0.00001$. The whitespace-excluded summary is negative with an
interval below zero in all three models, and so is the CAI measured directly as
one cell over all non-whitespace targets. The reference mixture is negative for
Qwen and Llama and positive for Gemma: at
Wikitext's $2.7\%$ whitespace share, the Gemma whitespace cell still contributes
$+0.019$, which outweighs the $-0.012$ from the other four types. The pooled
result therefore reflects the whitespace cell.

\begin{table}[t]
\centering
\small
\caption{Canonical GSM8K/Wikitext H=3 mean-blend top-1 audit on 300 sequences
per corpus: the pooled estimate, every conditional cell, and three summaries
(whitespace excluded, all non-whitespace targets measured as one cell, and the
reference mixture over all five cells), each with its 95\% sequence-bootstrap
interval. Under each model, the whitespace share of the evaluated targets in
GSM8K and in Wikitext. $n_\Ndom$ counts the Wikitext sequences that supply a
target of that type (Qwen/Llama/Gemma), and every cell uses all 300 GSM8K sequences.
The whitespace cell rests on 3,348/3,526/3,281 GSM8K and 1,093/1,171/1,150
Wikitext target tokens, and \tabref{tab:app_target_counts} gives all counts.}
\label{tab:canonical}
% Each interval sits on its own line under its estimate, so the table fits the
% text block at the caption's type size with the endpoints at four decimals.
\setlength{\tabcolsep}{3.5pt}
\begin{tabular}{@{}lcccr@{}}
\toprule
Estimate & Qwen2.5-7B & Llama-3.2-3B & gemma-2-2b-it & $n_\Ndom$ \\
 & ws $8.1\%$ / $2.5\%$ & ws $9.5\%$ / $2.8\%$ & ws $8.2\%$ / $2.7\%$ & \\
\midrule
pooled & $+0.0253$ & $+0.0428$ & $+0.0421$ & 300 \\
 & [$0.0211,0.0292$] & [$0.0389,0.0468$] & [$0.0376,0.0460$] & \\
\midrule
whitespace & $+0.5181$ & $+0.8215$ & $+0.7155$ & 217/217/218 \\
 & [$0.4786,0.5560$] & [$0.7920,0.8538$] & [$0.6879,0.7420$] & \\
content word & $-0.0077$ & $-0.0076$ & $-0.0019$ & 300 \\
 & [$-0.0097,-0.0056$] & [$-0.0099,-0.0051$] & [$-0.0040,0.0000$] & \\
number & $-0.0181$ & $-0.0055$ & $-0.0246$ & 219/218/220 \\
 & [$-0.0233,-0.0129$] & [$-0.0082,-0.0032$] & [$-0.0313,-0.0191$] & \\
punctuation & $-0.0981$ & $-0.1975$ & $-0.0350$ & 300 \\
 & [$-0.1099,-0.0867$] & [$-0.2125,-0.1821$] & [$-0.0482,-0.0202$] & \\
function word & $-0.0257$ & $-0.0138$ & $-0.0154$ & 300 \\
 & [$-0.0324,-0.0194$] & [$-0.0191,-0.0089$] & [$-0.0211,-0.0094$] & \\
\midrule
whitespace excluded & $-0.0247$ & $-0.0330$ & $-0.0131$ & 300 \\
 & [$-0.0272,-0.0220$] & [$-0.0355,-0.0303$] & [$-0.0157,-0.0104$] & \\
non-whitespace, measured & $-0.0253$ & $-0.0329$ & $-0.0155$ & 300 \\
 & [$-0.0280,-0.0227$] & [$-0.0355,-0.0303$] & [$-0.0186,-0.0125$] & \\
reference mixture & $-0.0121$ & $-0.0116$ & $+0.0074$ & 300 \\
 & [$-0.0154,-0.0087$] & [$-0.0158,-0.0073$] & [$0.0039,0.0111$] & \\
\bottomrule
\end{tabular}
\end{table}

\begin{figure}[t]
\centering
\includegraphics[width=\linewidth]{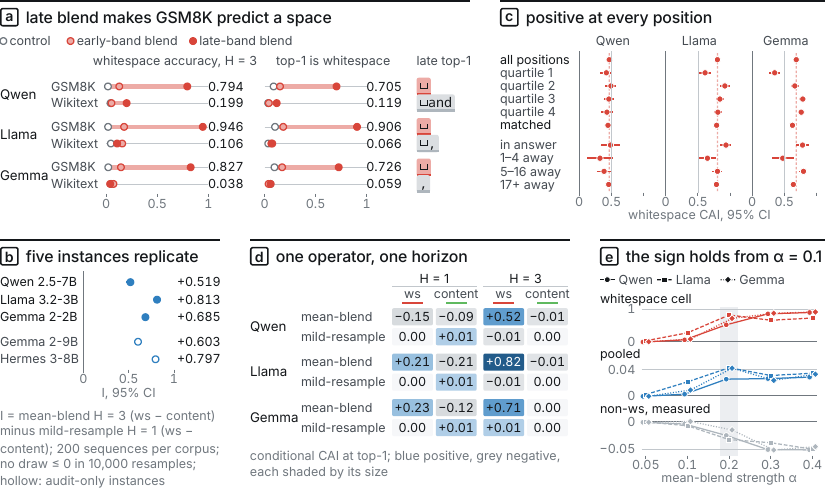}
\caption{The positive contrast arises from late-band mean-blend on GSM8K
whitespace. It persists across instances, positions, and strengths from
$\alpha=0.1$ to $0.4$, and its whitespace concentration is largest under
mean-blend three tokens ahead. Panel \textbf{a} gives H=3 whitespace
accuracy, averaged within sequences, and the share of positions whose top-1
prediction is whitespace, under control and the two blended bands. The chip is
the modal late-band top-1 token. Panel \textbf{b} gives the diagnostic interaction
$I$ of Eq.~\eqref{eq:audit} in the five same-sequence audit instances. Panel
\textbf{c} gives the whitespace CAI of the position-control sample by relative-position
quartile, after position matching, and by distance from the final answer span. The dashed line is each model's all-positions estimate. Panel \textbf{d}
gives the operator $\times$ horizon grid, and panel \textbf{e} the whitespace cell,
the pooled CAI and the directly measured non-whitespace CAI at five mean-blend
strengths, with $\alpha=0.2$ shaded. Intervals are 95\%. Appendix
Tables~\ref{tab:app_mechanism}, \ref{tab:app_prediction_probe},
\ref{tab:app_operator_horizon_grid}, \ref{tab:app_alpha_meanblend} and~\ref{tab:position_main}
list the values.}
\label{fig:mechanism}
\end{figure}

\paragraph{Where the positive whitespace contrast arises.}
On the canonical sample, late-band mean-blend raises H=3 whitespace accuracy on
GSM8K from about $0.02$ under control to between $0.79$ and $0.95$ in the three
models, and it changes Wikitext whitespace accuracy by at most $0.16$
(\figref{fig:mechanism}a, \tabref{tab:app_mechanism}). Split into its early and late parts, the whitespace
cell has a positive late part in every model, and its early part is negative, so
the late part exceeds the whole cell. The positive contrast therefore arises
from larger late-band accuracy gains on GSM8K whitespace. The predictions show
why: the late band makes a single space the top-1 prediction at $0.70$ to $0.91$
of GSM8K positions, whatever the target, and a space is the most frequent GSM8K
H=3 target. On Wikitext the late-band default is a comma or
\texttt{\textvisiblespace and} (\tabref{tab:app_prediction_probe}).

\paragraph{Replication and position.}
A separate 200-sequence audit measures both operators on the same sequences
(Appendix~\ref{app:audit}). Its diagnostic interaction $I$ is positive in all
five instances, from $+0.52$ to $+0.81$, with no non-positive draw among 10,000
bootstrap resamples (\figref{fig:mechanism}b). A position-control sample of 200 sequences per corpus then
evaluates the whitespace cell within relative-position quartiles and within bins
defined by each whitespace target's distance from the final GSM8K answer span.
Every quartile remains positive for every model. The position-matched estimate,
which reweights GSM8K to Wikitext's whitespace-position distribution, is also
positive in all three models. Each model's effect is positive both inside the
final answer clause and at least 17 tokens away from it, so the effect extends
beyond the phrase ``The answer is'' and the final sequence quartile
(\figref{fig:mechanism}c, \tabref{tab:position_main}).

\section{Screening corpus pairs before a model runs}
\label{sec:coverage}

Decomposing a CAI requires model runs on every corpus pair. A screen can flag a
pair in advance: it assigns a risk label from tokenizer-only means over the three
tokenizers, and each label predicts the region in which the pair's model outputs
will fall. \tabref{tab:screen_rule} gives both halves: the input rule, applied in
order until one line holds, and the observed region against which each label is
scored. JS is the Jensen--Shannon divergence between the two corpora's
target-type distributions. The input thresholds come from eleven calibration
rows, one per model instance on GSM8K/Wikitext, code/Wikitext, and HotpotQA/AGNews
(Appendix~\ref{app:predictive}). The pair selection and the input rule were fixed
before any intervention outcome was computed. The three output thresholds were set
with the calibration rows in view, which makes the held-out evaluation against
them exploratory. The scored assignments stay unchanged over a wide range of
cutoffs (Appendix~\ref{app:predictive}).

\begin{table}[H]
\centering
\small
\caption{The tokenizer-only screen. The input rule assigns the first label whose
condition holds, and each label is scored against the observed region, which is
computed from model outputs. ``Whitespace largest'' means that the H=3 mean-blend
whitespace cell exceeds both zero and the content-word cell.}
\label{tab:screen_rule}
\setlength{\tabcolsep}{4pt}
\begin{tabular}{@{}lL{0.43\linewidth}L{0.36\linewidth}@{}}
\toprule
Label & Input rule (tokenizer only) & Observed region and interval criterion \\
\midrule
boundary & focal corpus is source code, and either $\DeltaWS>0.04$ or newline
and indentation tokens exceed $25\%$ of its whitespace targets & not scored \\
high & $\DeltaWS\geq0.045$, $\DeltaChars\geq0.50$, and JS $\geq0.01$ &
$I\geq0.35$ with whitespace largest; $I$ interval above zero \\
low & $|\DeltaWS|\leq0.020$ and $|\DeltaChars|\leq0.35$ & $|I|\leq0.20$ and
$|\cai_{\rm pooled}|\leq0.08$; $I$ interval contains zero \\
medium & $\DeltaWS\geq0.030$ or $\DeltaChars\geq0.35$ & no held-out pair \\
ambiguous & otherwise & not scored \\
\bottomrule
\end{tabular}
\end{table}

\begin{figure}[t]
\centering
\includegraphics[width=\linewidth]{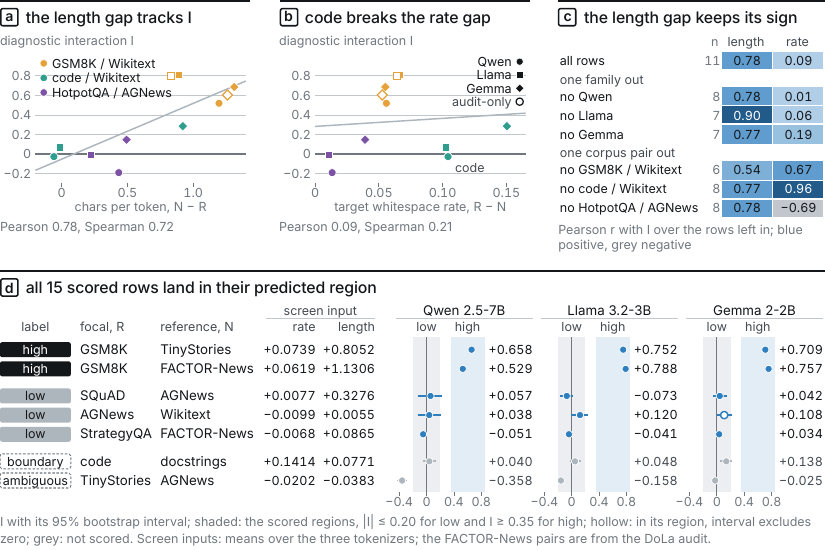}
\caption{The tokenizer-only screen and every pair it labelled in advance. Panels
\textbf{a} and \textbf{b} plot the diagnostic interaction $I$, measured from model
outputs, against the two tokenizer-only gaps for the eleven calibration rows:
characters per token, $\DeltaChars$, in \textbf{a} and target whitespace rate,
$\DeltaWS$, in \textbf{b}, each with its least-squares line. Panel \textbf{c}
gives both Pearson correlations over all eleven rows and with one tokenizer family
or one corpus pair left out. Panel \textbf{d} lists the five held-out pairs and
the two DoLa pairs with the screen's two inputs, the label fixed before any
intervention, and $I$ with its 95\% bootstrap interval in each primary model
(Tables~\ref{tab:app_heldout_outcomes} and~\ref{tab:app_dola}).}
\label{fig:screen}
\end{figure}

\paragraph{Held-out pairs.}
Five held-out pairs were run on the three primary models. Three received scored
labels (GSM8K/TinyStories high; SQuAD/AGNews and AGNews/Wikitext low), which gives
nine scored rows. Code against docstrings (boundary) and TinyStories/AGNews
(ambiguous) are reported unscored. All nine scored rows fall in the region their
label predicts, and eight of nine also meet the interval criterion. Gemma on
AGNews/Wikitext is the exception: its point estimates lie in the low region, and
its interval excludes zero (\tabref{tab:app_heldout_outcomes},
\figref{fig:screen}d).

\paragraph{DoLa corpus audit.}
We apply the CAI protocol to two corpus pairs built from benchmarks
evaluated by DoLa \citep{chuang2024dola}: GSM8K/FACTOR-News and
StrategyQA/FACTOR-News. The tokenizer-only screen rates these pairs
at different risk levels. FACTOR-News sequences are drawn from the
1,036-row test split of a public NEWS-FACTOR mirror
(Appendix~\ref{app:dola_update}). For high-risk GSM8K/FACTOR-News, the audit
interaction and pooled CAI are positive in every model, while both mixture
summaries are negative. For low-risk StrategyQA/FACTOR-News, every interaction
interval contains zero (\tabref{tab:app_dola}). All six prospective region
predictions agree with the observed outcomes. Our ablation contrast on
GSM8K/FACTOR-News therefore depends on the target mixture in the same way as the
canonical contrast.

\paragraph{Coverage and reversals.}
\tabref{tab:coverage} lists every decomposed pair with its per-model pooled and
whitespace-excluded estimates, and the reversals follow the screen's labels.
Among the pooled-positive rows of high- and medium-risk pairs, 14 of 14 reverse
sign once whitespace is excluded, each with a pooled interval wholly above zero
and a whitespace-excluded interval wholly below it
(Tables~\ref{tab:canonical} and~\ref{tab:app_expansion_intervals}). Reference
reweighting turns 11 of the same 14 rows negative (\tabref{tab:app_reversals}).
Among the other pooled-positive rows, 2 of 10 reverse, both from Gemma:
code/docstrings, with both intervals excluding zero, and SQuAD/AGNews, whose
pooled and whitespace-excluded intervals both contain zero. Within the reasoning
corpora, MATH and DROP have positive diagnostic interactions with intervals
excluding zero in every model row, and all ARC intervals include zero
(\tabref{tab:app_reasoning_extension}). Across the canonical, MATH, DROP, and
ARC rows, the whitespace-rate gap and the pooled CAI agree at Spearman
$\rho=0.937$.

\begin{table}[t]
\centering
\small
\caption{H=3 mean-blend top-1 CAI for every decomposed corpus pair, as pooled and
whitespace-excluded estimates per model. ``reversed'' counts rows whose
whitespace-excluded estimate is negative among rows with a positive pooled
estimate. $n$ is the number of sequences per corpus for Qwen/Llama/Gemma, and a value
that rounds to zero is printed to its first significant digit.
\tabref{tab:canonical} and \tabref{tab:app_expansion_intervals} give the 95\%
intervals, and
Appendix~\ref{app:coverage_accounting} lists two further screen-only pairs and the
condition count.}
\label{tab:coverage}
\setlength{\tabcolsep}{3pt}
\resizebox{\linewidth}{!}{%
\begin{tabular}{@{}lcrcccc@{}}
\toprule
 & & & \multicolumn{3}{c}{pooled / whitespace excluded} & \\
\cmidrule(lr){4-6}
Pair & screen & $n$ & Qwen2.5-7B & Llama-3.2-3B & gemma-2-2b-it & reversed \\
\midrule
GSM8K/Wikitext & high (cal.) & 300 & $+0.025$ / $-0.025$ & $+0.043$ / $-0.033$ & $+0.042$ / $-0.013$ & 3/3 \\
GSM8K/TinyStories & high & 200/195/200 & $+0.025$ / $-0.029$ & $+0.041$ / $-0.028$ & $+0.028$ / $-0.024$ & 3/3 \\
GSM8K/FACTOR-News & high & 200 & $+0.034$ / $-0.018$ & $+0.047$ / $-0.027$ & $+0.035$ / $-0.021$ & 3/3 \\
MATH/Wikitext & medium & 200 & $-0.009$ / $-0.024$ & $+0.006$ / $-0.031$ & $+0.012$ / $-0.008$ & 2/2 \\
DROP/Wikitext & medium & 168/185/159 & $+0.008$ / $-0.021$ & $+0.018$ / $-0.028$ & $+0.012$ / $-0.014$ & 3/3 \\
code/docstrings & boundary & 200 & $-0.005$ / $-0.019$ & $-0.006$ / $-0.013$ & $+0.017$ / $-0.035$ & 1/1 \\
AGNews/Wikitext & low & 82/75/82 & $-0.019$ / $-0.017$ & $-0.001$ / $+0.00003$ & $+0.005$ / $+0.005$ & 0/1 \\
SQuAD/AGNews & low & 118/119/138 & $+0.025$ / $+0.022$ & $-0.002$ / $-0.012$ & $+0.00002$ / $-0.008$ & 1/2 \\
ARC/Wikitext & low & 189/196/190 & $+0.004$ / $+0.005$ & $-0.009$ / $-0.024$ & $+0.007$ / $+0.007$ & 0/2 \\
StrategyQA/FACTOR-News & low & 150/135/149 & $-0.005$ / $-0.006$ & $-0.007$ / $-0.008$ & $+0.001$ / $+0.001$ & 0/1 \\
TinyStories/AGNews & ambiguous & 200 & $+0.010$ / $+0.021$ & $+0.006$ / $+0.006$ & $+0.006$ / $+0.005$ & 0/3 \\
\bottomrule
\end{tabular}}
\end{table}

\section{Robustness across definitions and readouts}
\label{sec:robustness}

The checks in this section examine how the findings vary with the
intervention, prediction horizon, token classification, and readout.

\paragraph{Operator and horizon.}
Within the operator-by-horizon grid, mean-blend H=3 is concentrated on
whitespace, while mild-resample H=3 is near zero for both target types.
Mild-resample H=1 gives a small positive content cell, which we split by
control accuracy below. The grid cells come from samples of different sizes, so
we compare them one cell at a time (\figref{fig:mechanism}d,
\tabref{tab:app_operator_horizon_grid}).

\paragraph{Classifier boundaries.}
We evaluate the primary horizon with four fixed vocabulary policies,
including two that move selected content tokens into the function-word
class and an adversarial map that assigns every token beginning with
whitespace to the whitespace class. That map dilutes the
whitespace effect by a factor of 8.2 to 8.7, and whitespace still holds the
largest positive H=3 cell in every model and the only positive cell in two. Among
the policy-conditioned non-whitespace estimates, only the Gemma content cell is
positive at H=3, under two policies, and both of its intervals include
zero. At H=1 under mean-blend, restricting the content class to tokens
without leading whitespace yields positive content-word CAI in
all three models (Appendix~\ref{app:classifier_policy}).

\paragraph{Continuous target margin.}
Under mean-blend at both H=1 and H=3, the target-margin CAI shows the same
selectivity. The six whitespace estimates run from $+1.23$ to $+3.37$ logits
with every interval excluding zero. Five of the six content estimates
are negative, and the remaining one, Gemma at H=1, is positive with an interval
that includes zero (\tabref{tab:app_target_margin}).
In three further checks (Appendix~\ref{app:beyond_cai_checks}),
log probability at H=3 gives a negative content-word CAI in every
primary model, the logit lens without ablation puts every H=3
late-band contrast near zero, and same-corpus H=1 patching stays
positive for content targets (\figref{fig:scope}c--d).
The positive pooled H=3 estimate also recurs
in both halves of an easy/hard GSM8K split, in three base checkpoints, and after
equalizing perturbation norms across corpora, and these controls pool over target
types (Appendix~\ref{app:controls}).

\paragraph{The one-token-ahead content-word cell.}
\label{sec:surviving}
In the operator grid, the only positive content cell lies at H=1 under
mild-resample. A separate H=1 mild-resample series of 250 sequences per corpus
measures it at top-1 and adds OLMo 2 \citep{olmo2024olmo2} and Qwen 3
\citep{qwen2025qwen3} to the three primary models. Its content-word CAI runs from $+0.0086$
to $+0.0148$ across the five instances, and its 95\% bootstrap interval excludes
zero in each (\tabref{tab:app_residual_support}). In every instance the unablated
model predicts a larger share of the GSM8K content targets than of the Wikitext
ones. The operator removes most of those correct predictions in both corpora,
more under the early band than under the late band. Each corpus's early-minus-late
damage gap therefore grows with its control accuracy. We split the content
targets by control-pass correctness and weight the two strata by their Wikitext
shares. The resulting estimates run from $-0.0028$ to $+0.0031$ with every
interval including zero, and the paired difference from the content cell excludes
zero in all five instances (\tabref{tab:app_h1_control_split}). The content cell
thus follows a second composition difference in the evaluated targets, the share
that the model already predicts. Two smaller effects remain: Llama keeps a
control-correct cell whose interval lies above zero, and reweighting by strata
that depend on the input alone, such as repeated tokens, leaves a smaller
positive cell (Appendix~\ref{app:h1_control_split}). For number, punctuation, and
function-word targets, three of the 15 reweighted intervals lie wholly above
zero, against eight unsplit cells above zero (\tabref{tab:app_h1_types}).

\section{Related work}
\label{sec:related}

Classical aggregation bias explains how a population mixture alters a pooled
contrast \citep{simpson1951}, and target-token composition is a concrete source
of it in cross-corpus depth attribution. In our case the whitespace conditional
effect is positive, so the pooled sign agrees with the stratum that has the
largest positive conditional effect, and the decomposition changes the task
interpretation attached to the pooled contrast. \tabref{tab:taxonomy}, in
Appendix~\ref{app:prior_patterns}, lists the reporting patterns most exposed to
this risk beside the evidence each one needs. Because the taxonomy concerns how
outputs are aggregated, it applies to any setting that pools scores over a
naturalistic target population, including layerwise readouts
\citep{nostalgebraist2020logitlens,belrose2023tunedlens,ghandeharioun2024patchscopes},
contrastive layer decoding \citep{chuang2024dola}, and causal tracing
\citep{meng2022rome}.

Ablation-validity studies vary interventions or faithfulness metrics
\citep{zhang2024activationpatching,li2024optimalablation,miller2024faithfulness}.
Our analysis complements these studies and analyses of attention structure and
expressivity \citep{lu2025linear,lu2026fem,lu2026hypermlp,lu2025zeros} by partitioning
targets under a fixed intervention, which concentrates the replicated H=3 contrast
on whitespace and leaves all four non-whitespace conditional point estimates
non-positive. Causal abstraction provides a broader framework for
intervention-based explanations \citep{geiger2025causalabstraction}.

\section{Discussion and limitations}
\label{sec:discussion}

A depth effect can replicate because its target-token mixture replicates, and in
our main case the replicated whitespace mixture carries the positive pooled
contrast. Across the high- and medium-risk rows with positive pooled
estimates, whitespace exclusion reverses every one.

\paragraph{What changes in practice.}
Report each corpus's target-type proportions and characters per token as a
low-cost tokenizer-only screen. Log the target-token ID beside each correctness
bit and report conditional cells: the decomposition and mixture checks reuse
predictions the experiment already collected. When a positive cell remains,
split its targets by whether the unablated model predicts them. The operator
mostly removes correct predictions, so a double difference of damage grows with
control accuracy.

\paragraph{Reading the two mixture summaries.}
Whitespace removal asks whether any broad non-whitespace effect remains, and it
is the decisive check in our high- and medium-risk rows. Reference reweighting
holds the measured conditional cells fixed and changes only their weights, so a
retained positive whitespace contribution can outweigh the negative
contributions of the other types, as it does for Gemma. The first summary
therefore tests for a non-whitespace effect, and the second shows how the
measured cells combine under the reference corpus's mixture. Whitespace can also
carry structure in code, markup, and tables, which is why the screen treats
source code as a boundary case.

\paragraph{Limitations.}
The primary claim concerns H=3 mean-blend top-1 CAI, and the continuous
target-margin readout corroborates it. The position analysis
covers canonical-pair whitespace, and the control-accuracy split covers the H=1
mild-resample series, so the positive H=1 mean-blend content cells of the
classifier-policy check remain unsplit. The main estimates use the perturbation
strength $\alpha=0.2$. In the strength sweep, the H=3 mean-blend whitespace
interval lies above zero in every model at every strength above the weakest, and
the directly measured non-whitespace interval never lies above zero
(\figref{fig:mechanism}e). The pooled H=1 mild-resample CAI interval lies wholly
above zero at most of the seven strengths in each model
(Appendix~\ref{app:alpha}). The
same protocol applies to any cross-corpus depth claim: report which tokens were
predicted, decompose the effect by target type, and match the mixture before
drawing a conclusion about depth.

\subsection*{AI use statement}

We used generative AI tools to implement the measurement, analysis, and figure
code, to reformat the public corpora into the evaluated prompt format, and to
interpret results. We
also used them to design the experiments, to give feedback on the methodology, and
to propose and refine the hypotheses that the runs were built to test. Finally, we
used them to develop the audit framework and to formulate its definitions. Those
definitions include the interaction estimand in \eqref{eq:cai}, the conditional
and mixture summaries, and the screening quantity in \eqref{eq:audit}. Synthetic data
generation, translation, mathematical proof, and qualitative analysis are not
applicable to this work, which evaluates the public corpora in their released
language, states no theorem, and analyzes no interview, survey, or free-text data.
We also used generative AI tools to create and edit code, draw figures, draft
and edit sections of this paper and its appendix, identify and summarize related
work, format references, and suggest the title and section structure.

We have reviewed all AI-assisted work. An internal verifier recomputes the
estimates, intervals, table cells, result counts, and ranges printed in the text
and tables of this paper and its appendix from the result artifacts, and it fails
when a printed result differs from the artifact that produced it. The
reported target-type decomposition, position controls, and held-out outcomes are
read from those artifacts. Prose descriptions of the intervention were checked
against the runner code that performs it, and the risk labels in
\secref{sec:coverage} were assigned before the held-out pairs were evaluated. We
take responsibility for the final content of this work, including text, claims,
and artifacts produced with the aid of generative AI.

\subsection*{Ethics statement}

Depth-attribution claims feed model auditing and deployment decisions, so a
measurement error in this class of study can become a false safety or capability
claim about where a competence lives inside a network. The audit makes the
evaluated target population explicit and reports an interval with each
conditional cell. It also splits the remaining positive content cell by the
unablated model's correctness, which relates that cell to the share of targets
the model already predicts. This work evaluates public checkpoints on public
corpora and involves no human subjects or new data collection. This arXiv
submission includes no research scripts, machine-readable result files, or model
weights. Appendix~\ref{app:impact} discusses broader-impact considerations.

\subsection*{Reproducibility statement}

The measurement protocol is specified in \secref{sec:method} and
Appendix~\ref{app:protocol}. They give the estimands, the deterministic
target-token classifier, the band, horizon, and operator definitions, and the
sequence-level resampling convention behind every reported interval. The main
text and appendix report the experimental results, sample counts, and numerical
summaries. An internal CPU verifier was used to cross-check the reported results
against the underlying result files. The analysis and evaluation scripts,
machine-readable result files, and figure-generation scripts are not provided
with this arXiv submission. Appendix~\ref{app:repro} summarizes the analyses and
the resample counts that differ across them. The corpora and checkpoints are the
public releases cited in \secref{sec:method}.

\newpage
\bibliographystyle{iclr2027_conference}
\bibliography{references}

\newpage
\appendix
\section{Audit protocol and additional definitions}
\label{app:protocol}

This appendix gives the full protocol and detailed results that support the
main text. For each aggregate depth result we report (i) a target-token
decomposition, (ii) mixture diagnostics, (iii) comparisons across
interventions, horizons, and readouts, and (iv) the one-token-ahead
content-token cell used for calibration, split by control accuracy.

\subsection{Notation and estimands}

For a corpus $C\in\{\Rdom,\Ndom\}$, offset $k$ (equivalently, horizon $H=k$),
band $B\in\{E,L\}$, and perturbation operator $o$, the damage score is
\[
\damage^{o}_{B}(C,k)=\acc_{\mathrm{ctrl}}(C,k)-\acc^{o}_{\mathrm{ablate}}(B,C,k).
\]
The contextual ablation interaction is
\[
\cai^{o}_{k}=\big[\damage^{o}_{E}(\Rdom,k)-\damage^{o}_{E}(\Ndom,k)\big]-\big[\damage^{o}_{L}(\Rdom,k)-\damage^{o}_{L}(\Ndom,k)\big].
\]
A positive value means that the early-band perturbation produces more excess
damage on $\Rdom$ than on $\Ndom$, relative to the corresponding late-band
contrast. The difference-in-differences estimand has a task-level interpretation when the
two corpora present comparable target-token prediction tasks.

For a token type $\tau$, the conditional interaction is
\begin{align*}
\cai^{o}_{k,\tau}
&=
\big[\damage^{o}_{E}(\Rdom,k\mid g(x_{t+k})=\tau)
-\damage^{o}_{E}(\Ndom,k\mid g(x_{t+k})=\tau)\big]
\\&\quad-
\big[\damage^{o}_{L}(\Rdom,k\mid g(x_{t+k})=\tau)
-\damage^{o}_{L}(\Ndom,k\mid g(x_{t+k})=\tau)\big].
\end{align*}
Let $\pi_C(\tau)$ be the target-type distribution in corpus $C$. The
mixture summary under a target distribution $q$ is
\[
M(q)=\sum_{\tau}q(\tau)\cai^o_{k,\tau}.
\]
The two summaries in the main text are
\begin{align*}
M_{\neg\ws} &= \sum_{\tau\ne\ws}\frac{w(\tau)}{\sum_{\sigma\ne\ws}w(\sigma)}\cai^o_{k,\tau}, \\
M_{\Ndom} &= \sum_{\tau}\pi_{\Ndom}(\tau)\cai^o_{k,\tau},
\end{align*}
where $w(\tau)$ is a target count. The canonical table takes $w$ from the two
corpora's combined counts, and the expansion and held-out rows take it from the
focal corpus's counts. $M_{\neg\ws}$ asks whether the effect survives after
removing whitespace targets. $M_{\Ndom}$ asks whether it survives when the
target-type mixture is changed to match the reference corpus.

\begin{figure}[H]
\centering
\includegraphics[width=\linewidth]{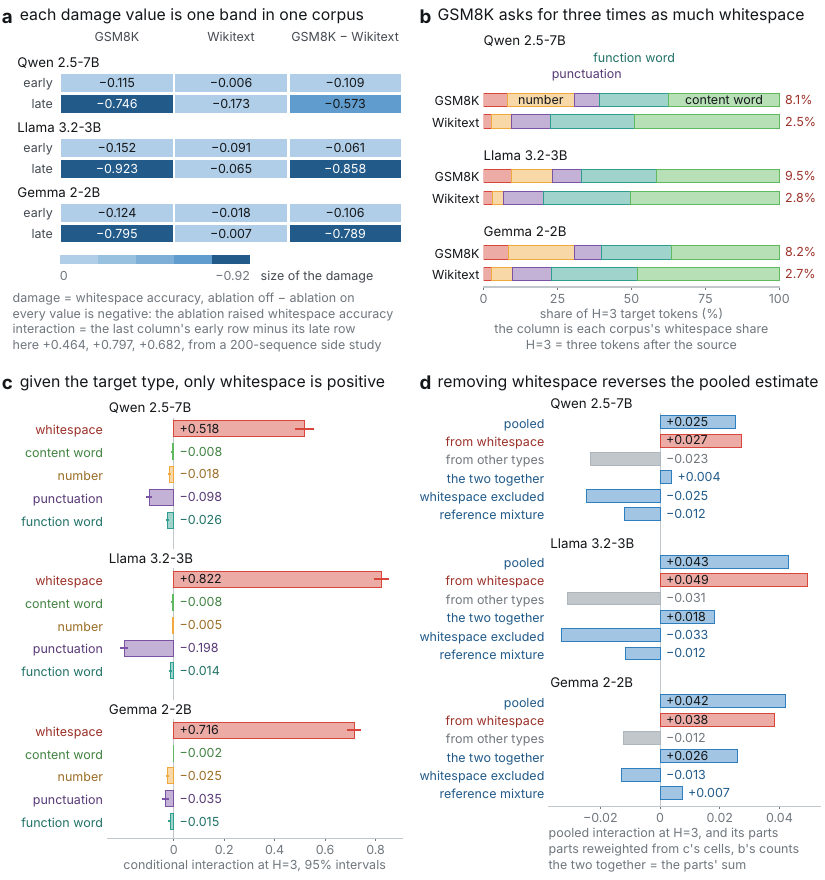}
\caption{The pooled depth interaction for GSM8K versus Wikitext is carried by one
target type. Panel \textbf{a} reports ablation damage by layer band, corpus, and
model for whitespace targets, on a separate 200-sequence sample. Its interaction
subtracts the late row of the last column from the early row. Panels
\textbf{b}--\textbf{d} come from the canonical 300-sequence sample. Panel
\textbf{b} shows each corpus's target-token mixture at a three-token-ahead
horizon (H=3), panel \textbf{c} gives the interaction conditional on target type
with 95\% intervals, and panel \textbf{d} places the pooled estimate beside
$M(q)$ at the combined target shares, split into its whitespace term and its
other four terms, and beside the two mixture summaries. \tabref{tab:canonical}
lists the values of panel \textbf{c} and the pooled and summary values of panel
\textbf{d} with their support, and Appendix~\ref{app:protocol} gives the three
$M(q)$ values.}
\label{fig:targettype}
\end{figure}

Both summaries reweight the measured conditional cells, and one identity relates
them to the pooled interaction. Write
$\Delta_C(\tau)=\damage^{o}_{E}(C,k\mid\tau)-\damage^{o}_{L}(C,k\mid\tau)$. If
accuracy is pooled over tokens, each pooled damage term is the mixture of its
conditional terms, so
\[
\cai^{o}_{k}=\sum_{\tau}\pi_{\Rdom}(\tau)\Delta_{\Rdom}(\tau)
            -\sum_{\tau}\pi_{\Ndom}(\tau)\Delta_{\Ndom}(\tau),
\qquad
M(q)=\sum_{\tau}q(\tau)\big[\Delta_{\Rdom}(\tau)-\Delta_{\Ndom}(\tau)\big],
\]
where $M(q)$ applies one mixture to both corpora. Taking $q=\pi_{\Rdom}$,
\[
\cai^{o}_{k}-M(\pi_{\Rdom})
=\sum_{\tau}\big[\pi_{\Rdom}(\tau)-\pi_{\Ndom}(\tau)\big]\Delta_{\Ndom}(\tau),
\]
which is zero when the two corpora share a target-type mixture. The estimator
averages accuracy within each sequence before averaging over sequences, so the
identity holds up to a residual. \tabref{tab:app_mixture} gives the pooled CAI,
$M(\pi_\Rdom)$, the mixture-shift term, and the residual for each model on the canonical sample. The residual is at most $0.0023$ in absolute value. The same
table splits $M(q)$ at the two corpora's combined target shares into its
whitespace term and its other four terms, which with their sum are the three
middle bars of each group in \figref{fig:targettype}d.

\subsection{Target-token classifier}

The classifier maps each vocabulary token to one of five categories by applying
the following rules, in order, to the token's decoded string after stripping
surrounding whitespace, and the first rule that matches assigns the category.
\begin{enumerate}[leftmargin=*,itemsep=0pt,topsep=2pt]
\item \emph{Whitespace}: the stripped string is empty, so the token consists of
spaces, tabs, newlines, or carriage returns.
\item \emph{Number}: every character is a digit or one of the period, comma, and
plus sign, and at least one character is a digit.
\item \emph{Punctuation}: every character is Unicode punctuation, a space, or a tab.
\item \emph{Function word}: the lower-cased string is one of 106 English
function words listed in the fixed inventory in Appendix~\ref{app:repro}.
\item \emph{Content word}: every remaining token, including any token that fails
to decode.
\end{enumerate}

This vocabulary-level map is applied to the target token $x_{t+k}$ at each
evaluated source position. Each class name reports the decoded string form of a vocabulary token. The map shows
whether mechanically distinct classes drive an aggregate metric. Mixture
diagnostics and prompt-level examples then test the resulting categories.

\subsection{Bands, horizons, and operators}

We partition layers into thirds, using the first third as the early band $E$
and the final third as the late band $L$. H=3 mean-blend measures the aggregate
artifact, and H=1 mild-resample gives the calibration setting and the
content-token cell of Appendix~\ref{app:h1_control_split}.
Both operators replace the output $h$ of every Transformer block in the band, at
every position, with
\[
h'=(1-\alpha)h+\alpha g,\qquad \alpha=0.2.
\]
Under mean-blend, $g$ is the mean of the same block's output at the same
position over the control pass of the same corpus. Under mild-resample, $g$ is
the same block's output at the same position in a donor sequence from the same
corpus. Donors are assigned by a derangement of the corpus's sequences, so no
sequence is its own donor, and positions past the donor's length are left
unchanged. The H=3 artifact is a mean-blend result, and the positive content-token cell is a mild-resample result. The diagnostic interaction in Eq.~\ref{eq:audit} takes the
whitespace-minus-content difference of conditional CAI under each setting,
mean-blend at H=3 and mild-resample at H=1, and subtracts the second from the
first.

\subsection{Model names}

Figure panels label the checkpoints in short form
(Qwen 2.5-7B, Qwen 3-1.7B, Llama 3.2-3B, Gemma 2-2B, Gemma 2-9B, Hermes 3-8B),
and the text and tables use the release identifiers, often without the
\texttt{-Instruct} or \texttt{-it} suffix. Outside \tabref{tab:app_base}, which
lists the base checkpoints, every model name denotes the instruction-tuned
checkpoint.

\subsection{Bootstrap and significance convention}

For comparisons across token types and intervention settings, each sequence is
the resampling unit. Unless otherwise stated, the main audit intervals use
10,000 sequence-level bootstrap resamples. The MATH, DROP, and ARC extensions
use 2,000 resamples, and other supplemental analyses state their own counts.
The one-sided $p$-value is the fraction of bootstrap draws for which the
interaction is non-positive. When no bootstrap draw is non-positive, the paper
reports the bound $p<10^{-4}$. Intervals and $p$-values are computed per estimand without multiplicity
adjustment.

\section{Prompt-level vignettes}
\label{app:vignettes}

The examples below show the token-type distinction directly. They were selected by readability and target type, using only the prompt window
and its target-token label.

\begin{table}[H]
\centering
\small
\caption{Prompt-level cases illustrating the target-token confound and the code
boundary. The bracketed token is the evaluated target.}
\label{tab:app_vignettes}
\begin{tabular}{@{}L{0.15\linewidth}L{0.14\linewidth}L{0.63\linewidth}@{}}
\toprule
Case & Target type & Context excerpt \\
\midrule
GSM8K H=3 & whitespace & \texttt{. That means the price for her ticket was 15 - 6 = \$9. The answer is[[[SPACE]]]9} \\
Wikitext H=3 & content & \texttt{12 Olympics . In July 2014 , she was featured in the ESPN 's The Body[[ Issue]] .} \\
Code H=3 & whitespace & \texttt{.X = vec1.X * vec2.X\quad new\_vec.Y = vec1.Y * vec2.Y[[\textbackslash n]]\quad return new\_vec} \\
GSM8K H=1 content cell & content & \texttt{supplies remaining in the ship is now 240-144 = 96 pounds. The[[ answer]] is 96} \\
\bottomrule
\end{tabular}
\end{table}

The first two rows show why an H=3 top-1 aggregate can compare different prediction
problems across domains. GSM8K frequently places a whitespace target before an
answer token, whereas the reference prose example places a content target. The
code row illustrates a boundary case in which structural whitespace motivates
reporting characters per token in addition to the target whitespace rate. The
final row shows a content-word target of the \(H=1\) mild-resample series.

\section{Complete target-token decomposition}
\label{app:decomposition}

\paragraph{Estimator provenance.}
Several tables in this appendix estimate the same \(H=3\) mean-blend whitespace
CAI cell using different sample sizes and analysis settings. The canonical
target-type decomposition, same-sequence bootstrap audit, and portability
analysis share the same estimand family and sign pattern. The analyses differ in sample composition, target-type support, missing-token
handling, bootstrap scope, and included readouts, so their values differ slightly.

\subsection{Conditional CAI cells}

\begin{table}[H]
\centering
\footnotesize
\caption{Canonical H=3 mean-blend target-token decomposition on GSM8K versus
Wikitext. Entries report CAI with 95\% bootstrap intervals in brackets.}
\label{tab:app_full_decomp}
\setlength{\tabcolsep}{2pt}
\begin{tabular}{@{}lccc@{}}
\toprule
Target type & Qwen2.5-7B & Llama-3.2-3B & gemma-2-2b-it \\
\midrule
whitespace & $+0.5181$ [$0.4786,0.5560$] & $+0.8215$ [$0.7920,0.8538$] & $+0.7155$ [$0.6879,0.7420$] \\
content & $-0.0077$ [$-0.0097,-0.0056$] & $-0.0076$ [$-0.0099,-0.0051$] & $-0.0019$ [$-0.0040,0.0000$] \\
number & $-0.0181$ [$-0.0233,-0.0129$] & $-0.0055$ [$-0.0082,-0.0032$] & $-0.0246$ [$-0.0313,-0.0191$] \\
punctuation & $-0.0981$ [$-0.1099,-0.0867$] & $-0.1975$ [$-0.2125,-0.1821$] & $-0.0350$ [$-0.0482,-0.0202$] \\
function word & $-0.0257$ [$-0.0324,-0.0194$] & $-0.0138$ [$-0.0191,-0.0089$] & $-0.0154$ [$-0.0211,-0.0094$] \\
\bottomrule
\end{tabular}
\end{table}

The sign structure carries the result. The positive effect is isolated to
whitespace targets, and content, number, punctuation, and function-word targets
are non-positive. The aggregate H=3 mean-blend effect therefore measures the
whitespace cell.

\subsection{Per-sequence records for the reweighted rows}

The intervals of the whitespace-excluded and reference-mixture rows of
\tabref{tab:canonical} need per-sequence records. A second evaluation pass with
the same data, length matching, 300 sequences per corpus, operator, and bands
keeps these records, and it reproduces the pooled CAI and all five conditional
cells of the decomposition to four decimals for all three models. Each of 2,000
bootstrap draws resamples sequences within each corpus, recomputes the five
cells, and recomputes the weights from the resampled target counts. The same records give the directly measured
non-whitespace row, which scores top-1 accuracy over all non-whitespace targets
of each sequence together. \tabref{tab:app_target_counts} gives the target
counts behind the weights and the support of the two cells that fall below 300
sequences.

\begin{table}[H]
\centering
\small
\caption{H=3 target counts in the canonical sample, 300 sequences per corpus.
The whitespace-excluded row of \tabref{tab:canonical} weights the four
non-whitespace cells by the summed counts of the two corpora, and the
reference-mixture row weights all five cells by the Wikitext ($\Ndom$) counts.
The last two columns count the sequences with at least one number or whitespace
target. Every other type appears in all 300 sequences of each corpus.}
\label{tab:app_target_counts}
\setlength{\tabcolsep}{4pt}
\begin{tabular}{llrrrrrrr}
\toprule
 & & \multicolumn{5}{c}{targets} & \multicolumn{2}{c}{sequences} \\
\cmidrule(lr){3-7}\cmidrule(lr){8-9}
Model & Corpus & content & number & punct. & whitespace & function & number & whitespace \\
\midrule
Qwen2.5-7B & GSM8K & 15{,}425 & 9{,}259 & 3{,}561 & 3{,}348 & 9{,}632 & 300 & 300 \\
Qwen2.5-7B & Wikitext & 21{,}193 & 2{,}884 & 5{,}696 & 1{,}093 & 12{,}245 & 219 & 217 \\
Llama-3.2-3B & GSM8K & 15{,}427 & 5{,}093 & 3{,}618 & 3{,}526 & 9{,}512 & 300 & 300 \\
Llama-3.2-3B & Wikitext & 20{,}786 & 1{,}609 & 5{,}603 & 1{,}171 & 12{,}051 & 218 & 217 \\
gemma-2-2b & GSM8K & 14{,}624 & 8{,}997 & 3{,}731 & 3{,}281 & 9{,}482 & 300 & 300 \\
gemma-2-2b & Wikitext & 20{,}639 & 3{,}049 & 5{,}610 & 1{,}150 & 12{,}610 & 220 & 218 \\
\bottomrule
\end{tabular}
\end{table}

\subsection{Weighting rule for the whitespace-excluded summary}

The canonical whitespace-excluded summary weights the four non-whitespace cells
by the combined target counts of the two corpora, renormalized to sum to one.
The expansion and held-out rows weight them by the focal corpus's counts. Under
the focal-corpus weights the canonical estimate is negative for all three models
(\tabref{tab:app_mixture}), so the reversal holds under either weighting rule.

\subsection{Target-type weights}

\begin{table}[H]
\centering
\small
\caption{Target-type distributions for GSM8K ($\Rdom$) and Wikitext ($\Ndom$) at H=3.}
\label{tab:app_weights}
\resizebox{\linewidth}{!}{%
\begin{tabular}{llrrrrr}
\toprule
Model & Domain & content & number & punctuation & whitespace & function-word \\
\midrule
Qwen2.5-7B-Instruct & GSM8K & 0.3742 & 0.2246 & 0.0864 & 0.0812 & 0.2336 \\
Qwen2.5-7B-Instruct & Wikitext & 0.4916 & 0.0669 & 0.1321 & 0.0254 & 0.2840 \\
Llama-3.2-3B-Instruct & GSM8K & 0.4150 & 0.1370 & 0.0973 & 0.0948 & 0.2559 \\
Llama-3.2-3B-Instruct & Wikitext & 0.5043 & 0.0390 & 0.1359 & 0.0284 & 0.2924 \\
gemma-2-2b-it & GSM8K & 0.3646 & 0.2243 & 0.0930 & 0.0818 & 0.2364 \\
gemma-2-2b-it & Wikitext & 0.4793 & 0.0708 & 0.1303 & 0.0267 & 0.2929 \\
\bottomrule
\end{tabular}}
\end{table}

At H=3, GSM8K has higher rates of whitespace and number targets than Wikitext.

\subsection{Mixture diagnostics}

\begin{table}[H]
\centering
\small
\caption{Mixture audit estimands for the canonical H=3 mean-blend interaction.
The first block gives the estimands that test whether the broad H=3
interpretation survives target-token exclusion and reference-mixture
reweighting. $\Delta_{\ws}$ is the
target whitespace-rate gap, a tokenizer-only quantity. The second block splits
the pooled CAI by the identity of Appendix~\ref{app:protocol}, on the per-sequence
records of Appendix~\ref{app:decomposition}, which reproduce the first row to four decimals. The third block splits
$M(q)$ at the combined target shares, as drawn in \figref{fig:targettype}d.}
\label{tab:app_mixture}
\begin{tabular}{lrrr}
\toprule
Estimand & Qwen2.5-7B-Instruct & Llama-3.2-3B-Instruct & gemma-2-2b-it \\
\midrule
raw pooled CAI & 0.0253 & 0.0428 & 0.0421 \\
$\Delta_{\ws}$-rate & 0.0559 & 0.0664 & 0.0551 \\
ws-excluded, combined weights & \textminus0.0247 & \textminus0.0330 & \textminus0.0131 \\
ws-excluded, focal weights & \textminus0.0233 & \textminus0.0294 & \textminus0.0143 \\
reference-reweighted & \textminus0.0121 & \textminus0.0116 & 0.0074 \\
\midrule
$M(\pi_\Rdom)$ & 0.0207 & 0.0513 & 0.0454 \\
mixture-shift term & 0.0057 & \textminus0.0097 & \textminus0.0011 \\
residual & \textminus0.0011 & 0.0012 & \textminus0.0023 \\
\midrule
whitespace term of $M(q)$ & 0.0273 & 0.0492 & 0.0381 \\
other four terms of $M(q)$ & \textminus0.0234 & \textminus0.0310 & \textminus0.0124 \\
$M(q)$ & 0.0039 & 0.0182 & 0.0257 \\
\bottomrule
\end{tabular}
\end{table}

The whitespace conditional
cell is positive, and every non-whitespace conditional cell in
Table~\ref{tab:app_full_decomp} is non-positive. The whitespace-excluded
estimate is negative for all three primary models, while reference-distribution
reweighting reduces the estimate to near zero or reverses its sign.

\section{Early/late mechanism decomposition}
\label{app:mechanism_update}

\tabref{tab:app_mechanism} gives the accuracy components behind the canonical
H=3 mean-blend whitespace cell, from the per-sequence records of
Appendix~\ref{app:decomposition}. Each accuracy is averaged within a sequence
and then over the sequences that have a whitespace target. Define the early
contribution as $D_E(\Rdom)-D_E(\Ndom)$ and the late contribution as
$D_L(\Ndom)-D_L(\Rdom)$, so their sum is the cell. Negative damage means that
the perturbation increased accuracy relative to control.

\begin{table}[H]
\centering
\small
\caption{Whitespace accuracy at H=3 under control, early-band, and late-band
mean-blend on the canonical sample, with the damage terms $D_E$ and $D_L$ and
the number $n$ of sequences with a whitespace target. The last two columns give
the early and late contributions to the whitespace cell of \tabref{tab:canonical}.}
\label{tab:app_mechanism}
\setlength{\tabcolsep}{4pt}
\begin{tabular}{llrrrrrrrr}
\toprule
Model & Corpus & $n$ & control & early & late & $D_E$ & $D_L$ & early contr. & late contr. \\
\midrule
Qwen2.5-7B & GSM8K & 300 & 0.0166 & 0.1273 & 0.7936 & $-0.1107$ & $-0.7770$ & $-0.1013$ & $+0.6194$ \\
 & Wikitext & 217 & 0.0418 & 0.0512 & 0.1994 & $-0.0094$ & $-0.1576$ & & \\
\addlinespace
Llama-3.2-3B & GSM8K & 300 & 0.0179 & 0.1747 & 0.9457 & $-0.1567$ & $-0.9277$ & $-0.0453$ & $+0.8667$ \\
 & Wikitext & 217 & 0.0451 & 0.1566 & 0.1061 & $-0.1114$ & $-0.0610$ & & \\
\addlinespace
gemma-2-2b & GSM8K & 300 & 0.0196 & 0.1438 & 0.8269 & $-0.1242$ & $-0.8073$ & $-0.1040$ & $+0.8194$ \\
 & Wikitext & 218 & 0.0504 & 0.0706 & 0.0383 & $-0.0202$ & $+0.0121$ & & \\
\bottomrule
\end{tabular}
\end{table}

The late contribution carries the whole positive cell in every model, because
late-band mean-blend lifts GSM8K whitespace accuracy from near zero to most
targets, a gain several times the size of its change on Wikitext in every model.
The early contribution is negative in every model.
\tabref{tab:app_prediction_probe} reads the predictions behind that gain. On
GSM8K the late band makes a single space the top-1 prediction at 0.70 to 0.91
of all positions, whereas whitespace is the top-1 prediction at 0.09 to 0.10 of
positions under control. It predicts whitespace at non-whitespace targets about
as often, so the prediction barely depends on the target type. A space is the most
frequent GSM8K H=3 target, at 0.08 to 0.09 of targets. On Wikitext the modal
late-band prediction is a comma or \texttt{\textvisiblespace and}, and
whitespace stays the top-1 prediction at no more than 0.12 of positions. The
late state thus moves predictions toward a frequent default continuation, and on
GSM8K that continuation matches many whitespace targets.

\begin{table}[H]
\centering
\small
\caption{Top-1 predictions behind \tabref{tab:app_mechanism}, from control and
late-band mean-blend passes at $\alpha=0.2$ on the same 300 sequences per corpus.
The columns give the share of all positions whose top-1 token is whitespace under
control and under the late band, the most frequent late-band top-1 token
(\texttt{\textvisiblespace} marks a space) and its share of positions, the share
of non-whitespace H=3 targets at which the late band predicts whitespace, and
the whitespace share of the H=3 targets.}
\label{tab:app_prediction_probe}
\setlength{\tabcolsep}{4pt}
\begin{tabular}{llrrlrrr}
\toprule
 & & \multicolumn{2}{c}{ws top-1} & \multicolumn{2}{c}{modal late top-1} & ws top-1 at & ws share \\
\cmidrule(lr){3-4}\cmidrule(lr){5-6}
Model & Corpus & control & late & token & share & non-ws targets & of targets \\
\midrule
Qwen2.5-7B & GSM8K & 0.088 & 0.705 & \texttt{\textvisiblespace{}} & 0.705 & 0.700 & 0.081 \\
 & Wikitext & 0.035 & 0.119 & \texttt{\textvisiblespace{}and} & 0.215 & 0.118 & 0.025 \\
\addlinespace
Llama-3.2-3B & GSM8K & 0.102 & 0.906 & \texttt{\textvisiblespace{}} & 0.906 & 0.900 & 0.095 \\
 & Wikitext & 0.036 & 0.066 & \texttt{\textvisiblespace{},} & 0.672 & 0.066 & 0.028 \\
\addlinespace
gemma-2-2b & GSM8K & 0.097 & 0.726 & \texttt{\textvisiblespace{}} & 0.726 & 0.717 & 0.082 \\
 & Wikitext & 0.056 & 0.059 & \texttt{,} & 0.371 & 0.060 & 0.027 \\
\bottomrule
\end{tabular}
\end{table}

\section{Position and answer-distance controls}
\label{app:position_control}

The position control applies the canonical H=3 position-aware mean-blend
protocol with $\alpha=0.2$ to 200 matched GSM8K and 200 matched Wikitext
sequences per model. \tabref{tab:app_position_bands} specifies the layer bands
for each architecture. Each whitespace target is annotated with its absolute
and relative sequence position, target ID, distance from GSM8K's final answer
span, and correctness under the control, early, and late conditions. We first
average accuracy within each sequence and then across sequences, with 95\%
intervals from 10,000 sequence-level bootstrap resamples drawn independently
within each domain.

\begin{figure}[H]
\centering
\includegraphics[width=\linewidth]{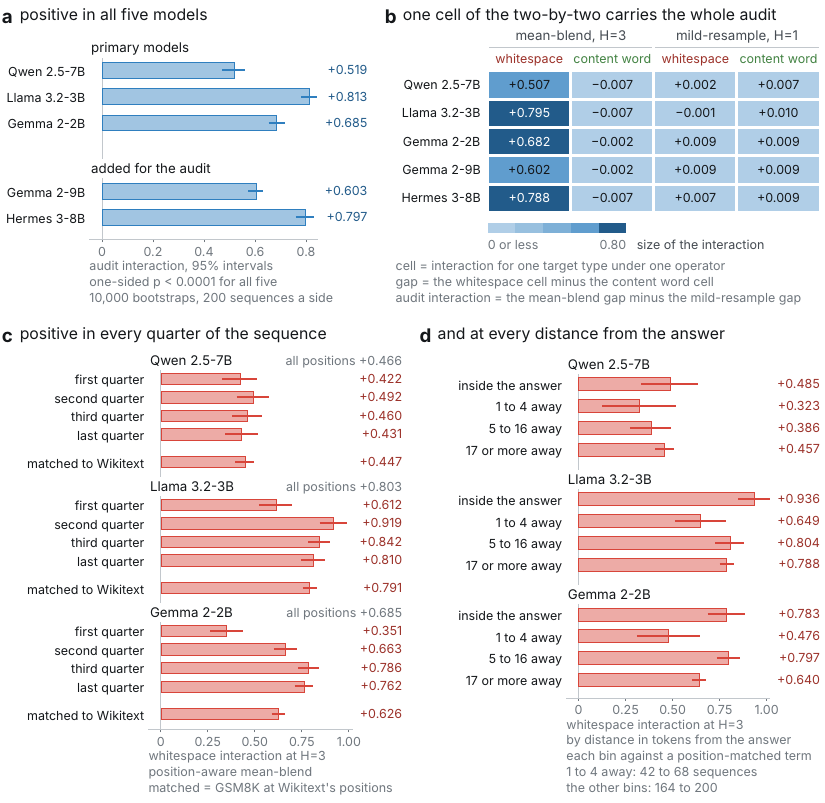}
\caption{The interaction persists across model instances and whitespace
positions. Panel \textbf{a} gives the diagnostic interaction per instance with
95\% intervals. Panel \textbf{b} shows whitespace and content-word CAI under each
operator and horizon for all five same-sequence audit instances. Panel
\textbf{c} gives the whitespace interaction by relative-position quartile and
after position reweighting. Panel \textbf{d} groups the same estimate by
distance from the final answer span, and \tabref{tab:app_answer_distance} gives
the event support.}
\label{fig:position}
\end{figure}

\begin{table}[H]
\centering
\scriptsize
\caption{Whitespace CAI by relative-position quartile. Event support is reported
as GSM8K/Wikitext. Sequence support varies because a sequence need not contain
a whitespace event in every quartile.}
\label{tab:app_position_quartiles}
\resizebox{\linewidth}{!}{%
\begin{tabular}{llrrr}
\toprule
Model & quartile & CAI & 95\% CI & events $\Rdom/\Ndom$ \\
\midrule
Qwen & Q1 & +0.422 & [0.327, 0.513] & 349/215 \\
 & Q2 & +0.492 & [0.403, 0.577] & 407/205 \\
 & Q3 & +0.460 & [0.380, 0.537] & 723/197 \\
 & Q4 & +0.431 & [0.341, 0.518] & 778/184 \\
\addlinespace
Llama & Q1 & +0.612 & [0.522, 0.696] & 371/225 \\
 & Q2 & +0.919 & [0.845, 0.992] & 370/194 \\
 & Q3 & +0.842 & [0.783, 0.900] & 788/189 \\
 & Q4 & +0.810 & [0.747, 0.871] & 850/204 \\
\addlinespace
Gemma & Q1 & +0.351 & [0.263, 0.437] & 359/228 \\
 & Q2 & +0.663 & [0.601, 0.725] & 397/221 \\
 & Q3 & +0.786 & [0.732, 0.840] & 706/198 \\
 & Q4 & +0.762 & [0.712, 0.811] & 761/184 \\
\addlinespace
\multicolumn{5}{l}{\scriptsize Full-sample events: Qwen 2,257/801, Llama 2,379/812, and Gemma 2,223/831.} \\
\bottomrule
\end{tabular}}
\end{table}

All 12 quartile intervals exclude zero, and the final table row gives the
full-sample event support. Quartile conditioning leaves a residual position
mixture, so we partition relative position into deciles and
reweight the GSM8K depth term to Wikitext's empirical whitespace-position
distribution.

\begin{table}[H]
\centering
\small
\caption{Position-matched whitespace estimate. The GSM8K depth term is
reweighted to Wikitext's relative-position decile weights. The Wikitext term
retains its empirical distribution.}
\label{tab:app_position_matched}
\begin{tabular}{lrr}
\toprule
Model & position-matched CAI & 95\% CI \\
\midrule
Qwen2.5-7B & +0.447 & [0.395, 0.496] \\
Llama-3.2-3B & +0.791 & [0.754, 0.828] \\
gemma-2-2b & +0.626 & [0.589, 0.663] \\
\bottomrule
\end{tabular}
\end{table}

For the answer-distance control, the final GSM8K answer span begins at the
last occurrence of the literal phrase \texttt{The answer is} and continues to
the end of the sequence. Each GSM8K bin is compared with a Wikitext depth
term reweighted to that bin's relative-position distribution, so each bin
contrast matches the two domains on relative position.

\begin{table}[H]
\centering
\scriptsize
\caption{GSM8K whitespace CAI by distance from the final answer span, each
compared with position-matched Wikitext. Support is GSM8K sequences/events.}
\label{tab:app_answer_distance}
\resizebox{\linewidth}{!}{%
\begin{tabular}{llrrr}
\toprule
Model & bin & CAI & 95\% CI & sequences/events \\
\midrule
Qwen & inside answer & +0.485 & [0.334, 0.635] & 200/200 \\
 & 1--4 tokens & +0.323 & [0.126, 0.520] & 46/46 \\
 & 5--16 tokens & +0.386 & [0.275, 0.492] & 169/331 \\
 & 17+ tokens & +0.457 & [0.407, 0.507] & 196/1680 \\
\addlinespace
Llama & inside answer & +0.936 & [0.848, 1.017] & 200/200 \\
 & 1--4 tokens & +0.649 & [0.512, 0.782] & 68/68 \\
 & 5--16 tokens & +0.804 & [0.726, 0.882] & 174/350 \\
 & 17+ tokens & +0.788 & [0.750, 0.827] & 198/1761 \\
\addlinespace
Gemma & inside answer & +0.783 & [0.686, 0.882] & 200/200 \\
 & 1--4 tokens & +0.476 & [0.310, 0.645] & 42/42 \\
 & 5--16 tokens & +0.797 & [0.735, 0.858] & 164/303 \\
 & 17+ tokens & +0.640 & [0.602, 0.677] & 196/1678 \\
\bottomrule
\end{tabular}}
\end{table}

Every answer-distance interval is positive, including the bin for targets at
least 17 tokens away, as \figref{fig:position}d shows.
\tabref{tab:app_answer_distance} gives each bin's sequence and event support.
The effect therefore survives conditioning on relative position and on distance
from the literal answer phrase.

\begin{table}[H]
\centering
\small
\caption{Layer bands used for the position control. Qwen and Llama use early
layers 0--8 and late layers 18--27 of 28 total layers.}
\label{tab:app_position_bands}
\begin{tabular}{lrrr}
\toprule
Model & total layers & early band & late band \\
\midrule
Qwen2.5-7B & 28 & 0--8 & 18--27 \\
Llama-3.2-3B & 28 & 0--8 & 18--27 \\
gemma-2-2b & 26 & 0--7 & 16--25 \\
\bottomrule
\end{tabular}
\end{table}

\section{Token type by intervention setting}
\label{app:audit}

\subsection{Audit cells}

\begin{table}[H]
\centering
\small
\caption{Same-sequence 2$\times$2 audit cells for
$I=(\mb_{H=3,\ws}-\mb_{H=3,\content})-(\rs_{H=1,\ws}-\rs_{H=1,\content})$.
The estimates use the audit sample and bootstrap analysis.}
\label{tab:app_audit_cells}
\resizebox{\linewidth}{!}{%
\begin{tabular}{lrrrrrr}
\toprule
Model & $\mb_{H=3,\ws}$ & $\mb_{H=3,\content}$ & $\rs_{H=1,\ws}$ & $\rs_{H=1,\content}$ & $I$ & 95\% CI \\
\midrule
Qwen2.5-7B-Instruct & 0.5066 & \textminus0.0074 & 0.0020 & 0.0072 & 0.5193 & [0.4705, 0.5621] \\
gemma-2-2b-it & 0.6820 & \textminus0.0022 & 0.0087 & 0.0094 & 0.6850 & [0.6534, 0.7192] \\
Llama-3.2-3B-Instruct & 0.7954 & \textminus0.0072 & \textminus0.0005 & 0.0102 & 0.8133 & [0.7811, 0.8436] \\
gemma-2-9b-it & 0.6015 & \textminus0.0019 & 0.0093 & 0.0093 & 0.6034 & [0.5729, 0.6325] \\
Hermes-3-Llama-3.1-8B & 0.7879 & \textminus0.0069 & 0.0069 & 0.0089 & 0.7968 & [0.7597, 0.8312] \\
\bottomrule
\end{tabular}}
\end{table}

\figref{fig:position}a gives each interaction with its 95\% interval, while
\figref{fig:position}b shows the same-sequence audit cells for all five
instances and \tabref{tab:app_audit_cells} reports those cells numerically. All
five audit interactions are positive, with no non-positive draw among 10,000
bootstrap resamples. In every instance the mean-blend H=3 whitespace cell
exceeds its content cell by more than 0.51. The mild-resample H=1 whitespace and
content cells agree to within 0.011.

\subsection{What the interaction subtracts}

A high whitespace CAI can come from a whitespace-sensitive domain feature or from
an operator artifact, and control accuracy differs across models. The interaction $I$ subtracts the content cell
within each operator and then subtracts the matched mild-resample difference,
so it measures how far the whitespace cell of the primary setting separates from
its content-token cell, relative to the same gap in the reference setting.

\section{Scope checks for the audit}
\label{app:p1_scope}

\subsection{Operator, horizon, and target-type grid}

\tabref{tab:app_operator_horizon_grid} expands the main audit contrast into a
compact operator $\times$ horizon $\times$ target-type grid on GSM8K/Wikitext.
The mean-blend and mild-resample cells come from estimates with different sample
sizes. The grid shows the sign pattern across the cells, and the main interaction gives the estimate with its bootstrap uncertainty.

\begin{table}[H]
\centering
\small
\caption{Operator $\times$ horizon $\times$ target-type grid on GSM8K/Wikitext.
Values are top-1 CAI\@. Mean-blend cells use the target-type decomposition,
whereas mild-resample cells use the effect-size analysis.}
\label{tab:app_operator_horizon_grid}
\resizebox{\linewidth}{!}{%
\begin{tabular}{llrrrr}
\toprule
Model & Operator & H=1 ws & H=1 content & H=3 ws & H=3 content \\
\midrule
Qwen2.5-7B & mean-blend & \textminus0.149 & \textminus0.093 & +0.518 & \textminus0.008 \\
Qwen2.5-7B & mild-resample & \textminus0.001 & +0.011 & \textminus0.013 & \textminus0.001 \\
Llama-3.2-3B & mean-blend & +0.208 & \textminus0.206 & +0.821 & \textminus0.008 \\
Llama-3.2-3B & mild-resample & \textminus0.004 & +0.012 & \textminus0.009 & \textminus0.0001 \\
gemma-2-2b & mean-blend & +0.231 & \textminus0.116 & +0.715 & \textminus0.002 \\
gemma-2-2b & mild-resample & \textminus0.004 & +0.009 & +0.014 & \textminus0.001 \\
\bottomrule
\end{tabular}}
\end{table}

The grid separates the two findings: whitespace concentration is largest for
mean-blend at H=3, whereas mild-resample H=3 is near zero for both target types.
Appendix~\ref{app:h1_control_split} splits the positive mild-resample content cell at H=1 by control accuracy.

\begin{figure}[H]
\centering
\includegraphics[width=\linewidth]{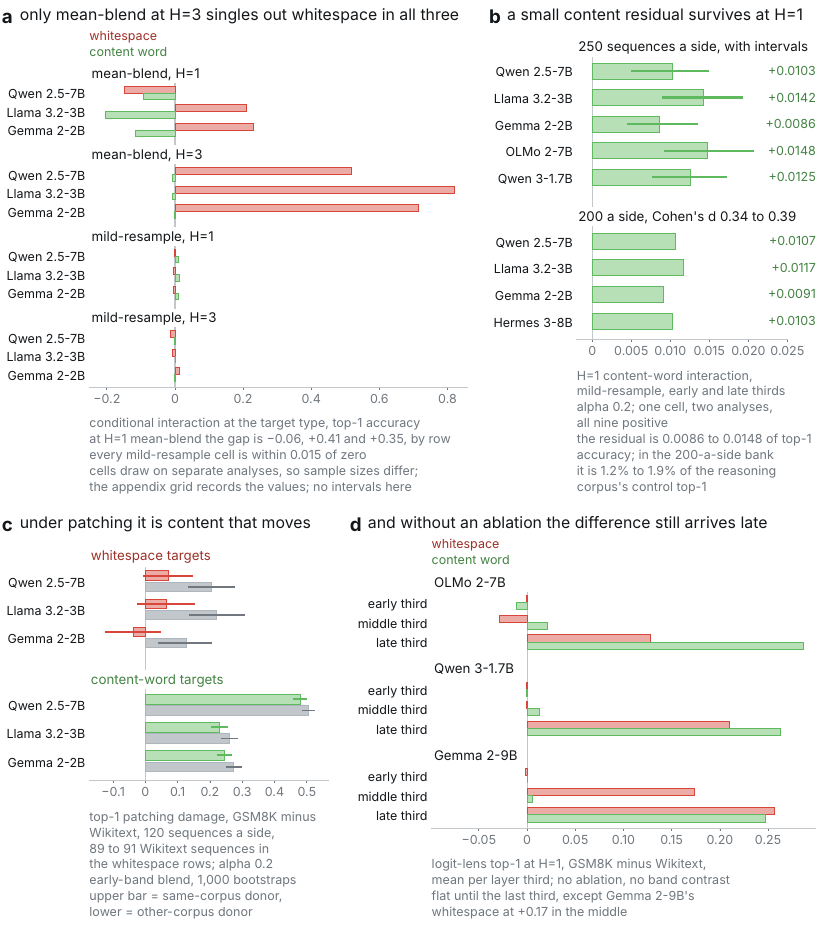}
\caption{Whitespace concentration is largest under one operator and horizon, and checks outside that cell yield small positive content-token effects. Panel \textbf{a}
shows whitespace and content-word CAI over the operator $\times$ horizon grid.
Panel \textbf{b} shows two measurements of the H=1 mild-resample
content-word CAI: 250 sequences per corpus with 95\% intervals, and 200 per corpus
with Cohen's $d$. Panel \textbf{c} gives the activation-patching check at top-1 by
target type, with the same-corpus donor above and the other-corpus donor below.
Panel
\textbf{d} reports unablated logit-lens top-1 accuracy at H=1, GSM8K minus Wikitext,
averaged within each layer third. \tabref{tab:app_operator_horizon_grid} records the grid cell by cell,
and Appendix~\ref{app:beyond_cai_checks} gives the patching and logit-lens
protocols.}
\label{fig:scope}
\end{figure}

\subsection{Checks beyond CAI}
\label{app:beyond_cai_checks}

These checks test whether target-type auditing remains informative
beyond top-1 CAI. For each readout or intervention, we ask which target-token
types drive the aggregate, whether mixture diagnostics change the conclusion,
and whether a task-appropriate non-whitespace cell remains.

\subsubsection{Log-probability CAI target-type audit}

\tabref{tab:app_beyond_cai_b1} compares target log probability with a matched
top-1 estimate. The top-1 artifact remains concentrated on whitespace in all
three primary models. The two readouts yield different results: the pooled
result reverses for Qwen, whereas Llama and Gemma remain positive before
reference reweighting. Under log probability, the content, punctuation, and function-word cells and the
reference-reweighted estimate are negative in all three models, and the number
cell is positive for Llama and Gemma.

\begin{table}[H]
\centering
\small
\caption{H=3 mean-blend CAI under top-1 and log-probability readouts, split by
target type on GSM8K/Wikitext. The top-1 rows provide a matched comparison for
log probability. ``ws-excl.''\@ renormalizes over non-whitespace types, and
``ref-rw.''\@ applies Wikitext target-type proportions.}
\label{tab:app_beyond_cai_b1}
\resizebox{\linewidth}{!}{%
\begin{tabular}{llrrrrrr}
\toprule
Model & Readout & pooled & ws & content & nonws & ws-excl. & ref-rw. \\
\midrule
Qwen2.5-7B & top-1 & +0.026 & +0.464 & \textminus0.008 & \textminus0.023 & \textminus0.021 & \textminus0.010 \\
Qwen2.5-7B & logprob & \textminus1.031 & +0.444 & \textminus1.062 & \textminus1.298 & \textminus1.222 & \textminus1.381 \\
Llama-3.2-3B & top-1 & +0.042 & +0.797 & \textminus0.008 & \textminus0.033 & \textminus0.030 & \textminus0.011 \\
Llama-3.2-3B & logprob & +0.141 & +1.448 & \textminus0.266 & \textminus0.114 & +0.056 & \textminus0.166 \\
gemma-2-2b & top-1 & +0.041 & +0.682 & \textminus0.002 & \textminus0.016 & \textminus0.015 & +0.007 \\
gemma-2-2b & logprob & +0.249 & +0.961 & \textminus0.046 & +0.097 & \textminus0.012 & \textminus0.147 \\
\bottomrule
\end{tabular}}
\end{table}

\subsubsection{Continuous target-margin audit}
\label{app:continuous_margin}

The target margin is the gold target token's logit minus the largest competing
logit. It changes continuously as a perturbation moves the gold token, including
movements that leave the gold token at rank one. The damage and CAI equations
keep their form, with mean target margin as the scored quantity. We retain conditioning
by horizon (H=1/H=3) and target type (whitespace/content).

\begin{table}[H]
\centering
\small
\caption{Target-margin CAI on GSM8K/Wikitext. Brackets are 95\% bootstrap
intervals.}
\label{tab:app_target_margin}
\resizebox{\linewidth}{!}{%
\begin{tabular}{lrrrr}
\toprule
Model & H=1 whitespace & H=1 content & H=3 whitespace & H=3 content \\
\midrule
Qwen2.5-7B & +1.661 [1.338, 1.974] & \textminus0.535 [\textminus0.758, \textminus0.313] & +1.721 [1.413, 2.035] & \textminus1.188 [\textminus1.354, \textminus1.013] \\
Llama-3.2-3B & +2.198 [1.971, 2.420] & \textminus0.819 [\textminus1.039, \textminus0.617] & +3.374 [3.101, 3.644] & \textminus0.388 [\textminus0.530, \textminus0.246] \\
gemma-2-2b & +1.229 [1.010, 1.444] & +0.172 [\textminus0.018, 0.368] & +1.593 [1.374, 1.803] & \textminus0.279 [\textminus0.420, \textminus0.145] \\
\bottomrule
\end{tabular}}
\end{table}

All six whitespace intervals exclude zero, while five of the six content point
estimates are negative and the remaining Gemma H=1 interval includes zero. The
H=3 target-type asymmetry therefore holds on a continuous scale as well, with
different CAI magnitudes on the two readouts.

\subsubsection{Non-CAI layerwise lens audit}

\tabref{tab:app_beyond_cai_b2} reports a non-intervention residual-stream
logit-lens readout on the same three primary models. The estimand is a layerwise
top-1 domain difference, averaged over the late third of layers and split by
target type. At H=1, the content-target late-layer contrast exceeds the pooled
estimate in all three models. At H=3, the late-layer contrasts are near zero, so the
CAI artifact is absent from this readout. Target-type conditioning therefore
changes the interpretation of this layerwise readout, which lies outside the CAI estimand.

\begin{table}[H]
\centering
\small
\caption{Non-CAI logit-lens target-type audit. Values are late-band top-1
domain differences on GSM8K/Wikitext, averaged over the final third of layers.}
\label{tab:app_beyond_cai_b2}
\resizebox{\linewidth}{!}{%
\begin{tabular}{lrrrrrrrr}
\toprule
Model & H=1 pooled & H=1 ws & H=1 cont. & H=1 nonws & H=3 pooled & H=3 ws & H=3 cont. & H=3 nonws \\
\midrule
Qwen2.5-7B & +0.1332 & +0.0778 & +0.1918 & +0.1365 & +0.0004 & \textminus0.0047 & +0.0012 & +0.0005 \\
Llama-3.2-3B & +0.1908 & +0.1961 & +0.2460 & +0.1928 & +0.0021 & \textminus0.0041 & +0.0051 & +0.0025 \\
gemma-2-2b & +0.1678 & +0.1004 & +0.1984 & +0.1657 & +0.0026 & \textminus0.0167 & +0.0020 & +0.0030 \\
\bottomrule
\end{tabular}}
\end{table}

\subsubsection{Activation patching with target-type uncertainty}
\label{app:beyond_cai_patching}

\tabref{tab:app_beyond_cai_b4} reports an H=1 activation-patching check with
target-type bootstrap intervals. Each estimate is GSM8K damage minus Wikitext
damage, where damage is the control H=1 score minus the score after each
early-band layer's output is blended, position by position, toward a donor
sequence's at $\alpha=0.2$. The
same-domain donor is another sequence of the same corpus, and the cross-domain
donor is the aligned sequence of the other corpus. In the table the score is
target log probability. Same-domain patching is positive for content and
non-whitespace targets in all three primary models. Cross-domain patching is
strongly reversed in Qwen, reversed or null in Gemma, and attenuated but still
weakly positive for non-whitespace targets in Llama. These results separate
same-domain from cross-domain patching, and the cross-domain sign varies by
model. \figref{fig:scope}c shows the same intervention scored by
top-1 correctness for whitespace and content-word targets.

\begin{table}[H]
\centering
\small
\caption{Activation-patching target-type check. Values are H=1 log-probability
effects with 95\% bootstrap intervals.}
\label{tab:app_beyond_cai_b4}
\resizebox{\linewidth}{!}{%
\begin{tabular}{lrrrr}
\toprule
Model & content same & content cross & nonws same & nonws cross \\
\midrule
Qwen2.5-7B & +9.725 [8.891, 10.576] & \textminus4.723 [\textminus5.510, \textminus3.842] & +8.632 [7.798, 9.365] & \textminus4.763 [\textminus5.515, \textminus3.949] \\
Llama-3.2-3B & +1.616 [1.093, 2.186] & +0.048 [\textminus0.483, 0.581] & +2.225 [1.772, 2.715] & +0.694 [0.149, 1.172] \\
gemma-2-2b & +1.654 [1.108, 2.178] & \textminus0.218 [\textminus0.784, 0.325] & +1.641 [1.183, 2.105] & \textminus0.170 [\textminus0.692, 0.299] \\
\bottomrule
\end{tabular}}
\end{table}

These results show that the target-type audit applies to aggregate readouts
and interventions over naturalistic domain contrasts.

\section{Prior-pattern taxonomy}
\label{app:prior_patterns}

The audit applies to reporting patterns that aggregate over naturalistic
target populations. Table~\ref{tab:taxonomy} summarizes the main cases and
our results for two comparisons motivated by DoLa. The extended taxonomy in
Table~\ref{tab:app_prior_patterns} separates these
aggregates from fixed-target settings, where domain-level target mixtures
pose less risk.

\begin{table}[H]
\centering
\scriptsize
\caption{At-risk reporting patterns and the evidence needed before a broad
depth claim. DoLa motivates the two corpus pairs measured here with our operators, and the other rows identify settings in which the diagnostic
applies.}
\label{tab:taxonomy}
\begin{tabular}{@{}L{0.18\linewidth}L{0.31\linewidth}L{0.41\linewidth}@{}}
\toprule
Pattern & Risk & Required check / result here \\
\midrule
Layerwise readout & Pooling next-token scores across naturalistic corpora & Target-type cells and mixture matching \\
Layer contrast / DoLa & Corpus-level layer buckets may covary with format & GSM8K/FACTOR high-risk in 3/3. StrategyQA/FACTOR low-region in 3/3 \\
Causal tracing / patching & Localization score depends on target population as well as intervention & Log target id and condition on task-appropriate targets \\
Ablation CAI & Replicated pooled interaction may inherit a stable mixture gap & Canonical pooled effect becomes negative after whitespace removal in 3/3 \\
\bottomrule
\end{tabular}
\end{table}

\begin{table}[H]
\centering
\small
\caption{Reporting patterns covered by the target-type audit.}
\label{tab:app_prior_patterns}
\begin{tabular}{p{0.28\linewidth}p{0.30\linewidth}p{0.32\linewidth}}
\toprule
Experimental pattern & Why it is at risk & What the audit changes \\
\midrule
Layerwise top-1 next-token readouts over math/prose or code/prose corpora, as in logit-lens, tuned-lens, or patchscope-style studies when summarized by naturalistic-domain accuracy & The readout pools different target-token populations before assigning a depth label & Report target-type cells and reweighted estimates before interpreting a replicated layer peak as depth specialization \\
Layer-contrast or intervention scores aggregated over naturalistic domain pairs, including DoLa-style layer contrasts or activation-patching variants when used as domain-level next-token metrics & The layer or intervention contrast can inherit tokenizer and target-mixture asymmetries from the benchmark & Surviving non-whitespace cells and mixture-matched estimates support the broad claim. Their collapse identifies a stable measurement confound \\
Fixed-target factual tracing or prompt-local causal restoration, such as ROME-style factual association localization & The target token is usually part of the constructed prompt or fact, so domain mixture is a secondary threat there & The audit becomes relevant when the same tools are scaled to naturalistic domain aggregates \\
Scenario- or option-level evaluations with matched answer formats & Target-token composition is partly controlled by the benchmark design & Report the format control. Target-type auditing becomes important if results are converted into free-form next-token aggregates \\
\bottomrule
\end{tabular}
\end{table}

\section{Condition accounting for the coverage table}
\label{app:coverage_accounting}

\tabref{tab:coverage} reports one row per corpus pair. The table contains 33
decomposed model--pair rows across the eleven fully decomposed pairs. Two
further pairs enter only the calibration screen: code/Wikitext, labelled
boundary, and HotpotQA/AGNews, the lower-risk calibration region of
\tabref{tab:risk_regions}. The HotpotQA/AGNews mean gaps, $\DeltaWS=0.021$ and
$\DeltaChars=0.38$, meet the medium rule of \tabref{tab:screen_rule}. For
these pairs the audit measured
the diagnostic interaction $I$ on each primary model
(\tabref{tab:app_predictive_rows}) without a pooled or whitespace-excluded
decomposition, so they have no row in \tabref{tab:coverage}. The canonical pair
contributes two audit-only instances, and each screen-only pair adds three
instances. The total is therefore 41 distinct conditions over 13 pairs
and five instances once these additions are included.
\tabref{tab:app_operator_horizon_grid} records the two central interventions at
both horizons, and \tabref{tab:app_expansion_intervals} gives 95\% intervals for
the pooled and whitespace-excluded values of the ten expansion pairs, one row per
pair and model.

\begin{table}[H]
\centering
\footnotesize
\caption{H=3 mean-blend top-1 CAI for the ten expansion pairs of
\tabref{tab:coverage}, one row per pair and model (30 rows), pooled and with whitespace excluded, each with its 95\%
sequence-bootstrap interval. Each of 2,000 draws resamples sequences within each
corpus and recomputes the cells, with the weights held at the screen's target-type
shares. The intervals come from an evaluation pass that repeats each audit's
sequences, operator, bands, and horizon and reproduces every pooled value,
conditional cell, and mixture summary of the audit to within $10^{-6}$. The number cells of GSM8K/TinyStories and TinyStories/AGNews,
with 4 and 1 sequence pairs, fall below the support rule of 10 and stay out of
the whitespace-excluded summary, as in the audit, and
\tabref{tab:app_expansion_support} gives every row's support. Measured directly
over all non-whitespace targets, numbers included, the CAI of these six rows
has the sign of their whitespace-excluded values, from $-0.0314$ to $-0.0289$ on
GSM8K/TinyStories and from $+0.0064$ to $+0.0222$ on TinyStories/AGNews. A draw in which a retained cell falls below the rule is
dropped, which leaves 1,750 to 1,997 draws in five whitespace-excluded rows,
the three ARC rows and the Qwen and Llama SQuAD rows. A value that rounds to zero is printed to its first significant
digit.}
\label{tab:app_expansion_intervals}
\setlength{\tabcolsep}{3pt}
\begin{tabular}{@{}llrrcrc@{}}
\toprule
Pair & model & $n$ & pooled & 95\% interval & ws-excl. & 95\% interval \\
\midrule
GSM8K/TinyStories & Qwen & 200 & $+0.0254$ & [0.0209, 0.0298] & $-0.0291$ & [\textminus0.0321, \textminus0.0262] \\
 & Llama & 195 & $+0.0407$ & [0.0358, 0.0453] & $-0.0280$ & [\textminus0.0307, \textminus0.0255] \\
 & Gemma & 200 & $+0.0276$ & [0.0211, 0.0336] & $-0.0238$ & [\textminus0.0273, \textminus0.0201] \\
\addlinespace
GSM8K/FACTOR-News & Qwen & 200 & $+0.0340$ & [0.0297, 0.0386] & $-0.0182$ & [\textminus0.0213, \textminus0.0151] \\
 & Llama & 200 & $+0.0467$ & [0.0420, 0.0514] & $-0.0266$ & [\textminus0.0296, \textminus0.0236] \\
 & Gemma & 200 & $+0.0351$ & [0.0286, 0.0413] & $-0.0207$ & [\textminus0.0247, \textminus0.0167] \\
\addlinespace
MATH/Wikitext & Qwen & 200 & $-0.0087$ & [\textminus0.0133, \textminus0.0040] & $-0.0242$ & [\textminus0.0276, \textminus0.0208] \\
 & Llama & 200 & $+0.0062$ & [0.0007, 0.0120] & $-0.0314$ & [\textminus0.0358, \textminus0.0269] \\
 & Gemma & 200 & $+0.0120$ & [0.0072, 0.0167] & $-0.0079$ & [\textminus0.0112, \textminus0.0045] \\
\addlinespace
DROP/Wikitext & Qwen & 168 & $+0.0078$ & [0.0032, 0.0126] & $-0.0211$ & [\textminus0.0239, \textminus0.0180] \\
 & Llama & 185 & $+0.0179$ & [0.0136, 0.0222] & $-0.0283$ & [\textminus0.0311, \textminus0.0256] \\
 & Gemma & 159 & $+0.0123$ & [0.0085, 0.0164] & $-0.0143$ & [\textminus0.0171, \textminus0.0113] \\
\addlinespace
code/docstrings & Qwen & 200 & $-0.0051$ & [\textminus0.0099, \textminus0.0009] & $-0.0194$ & [\textminus0.0243, \textminus0.0150] \\
 & Llama & 200 & $-0.0060$ & [\textminus0.0101, \textminus0.0019] & $-0.0131$ & [\textminus0.0178, \textminus0.0087] \\
 & Gemma & 200 & $+0.0166$ & [0.0117, 0.0216] & $-0.0352$ & [\textminus0.0407, \textminus0.0294] \\
\addlinespace
AGNews/Wikitext & Qwen & 82 & $-0.0192$ & [\textminus0.0264, \textminus0.0119] & $-0.0174$ & [\textminus0.0250, \textminus0.0103] \\
 & Llama & 75 & $-0.0006$ & [\textminus0.0097, 0.0086] & $+0.00003$ & [\textminus0.0086, 0.0090] \\
 & Gemma & 82 & $+0.0051$ & [\textminus0.0021, 0.0122] & $+0.0050$ & [\textminus0.0032, 0.0131] \\
\addlinespace
SQuAD/AGNews & Qwen & 118 & $+0.0245$ & [0.0158, 0.0336] & $+0.0221$ & [0.0127, 0.0321] \\
 & Llama & 119 & $-0.0017$ & [\textminus0.0078, 0.0050] & $-0.0120$ & [\textminus0.0197, \textminus0.0041] \\
 & Gemma & 138 & $+0.00002$ & [\textminus0.0081, 0.0083] & $-0.0082$ & [\textminus0.0178, 0.0014] \\
\addlinespace
ARC/Wikitext & Qwen & 189 & $+0.0038$ & [\textminus0.0008, 0.0082] & $+0.0051$ & [0.0004, 0.0096] \\
 & Llama & 196 & $-0.0089$ & [\textminus0.0131, \textminus0.0047] & $-0.0241$ & [\textminus0.0293, \textminus0.0192] \\
 & Gemma & 190 & $+0.0074$ & [0.0039, 0.0109] & $+0.0070$ & [0.0033, 0.0108] \\
\addlinespace
StrategyQA/ & Qwen & 150 & $-0.0048$ & [\textminus0.0094, \textminus0.00002] & $-0.0055$ & [\textminus0.0102, \textminus0.0007] \\
FACTOR-News & Llama & 135 & $-0.0070$ & [\textminus0.0118, \textminus0.0026] & $-0.0075$ & [\textminus0.0121, \textminus0.0031] \\
 & Gemma & 149 & $+0.0009$ & [\textminus0.0052, 0.0071] & $+0.0011$ & [\textminus0.0047, 0.0072] \\
\addlinespace
TinyStories/AGNews & Qwen & 200 & $+0.0099$ & [0.0058, 0.0141] & $+0.0205$ & [0.0159, 0.0252] \\
 & Llama & 200 & $+0.0060$ & [0.0021, 0.0096] & $+0.0055$ & [0.0008, 0.0101] \\
 & Gemma & 200 & $+0.0060$ & [0.0012, 0.0108] & $+0.0050$ & [\textminus0.0004, 0.0101] \\
\bottomrule
\end{tabular}
\end{table}

\begin{table}[H]
\centering
\footnotesize
\caption{Support behind \tabref{tab:app_expansion_intervals}. A sequence pair is
the focal and the reference sequence at one index. Each sequence receives three
passes, the unablated control pass and the early-band and late-band mean-blend
passes, and a pair enters a cell when both sequences have an H=3 target of that
type in all three. The content, punctuation, and function-word cells use every
pair in every row except the code/docstrings function-word cells, at 193, 193,
and 192 of 200 for Qwen, Llama, and Gemma. Target counts are the H=3 targets of
each corpus in the control pass, focal / reference, and the non-whitespace column
sums the four other types. $^{*}$ marks a cell below the support rule of 10 pairs,
which is not estimated. The six starred number cells stay out of their rows'
whitespace-excluded summaries. The starred whitespace cell, Llama on SQuAD/AGNews,
enters neither printed value of \tabref{tab:app_expansion_intervals}, since the
pooled value uses all targets and the whitespace-excluded value drops whitespace.}
\label{tab:app_expansion_support}
\setlength{\tabcolsep}{3pt}
\begin{tabular}{@{}llrrrrrr@{}}
\toprule
 & & & \multicolumn{2}{c}{whitespace} & \multicolumn{2}{c}{number} & non-whitespace \\
\cmidrule(lr){4-5}\cmidrule(lr){6-7}
Pair & model & $n$ & pairs & targets & pairs & targets & targets \\
\midrule
GSM8K/TinyStories & Qwen & 200 & 89 & 2{,}600 / 208 & 4$^{*}$ & 7{,}002 / 4 & 28{,}837 / 33{,}433 \\
 & Llama & 195 & 84 & 2{,}722 / 192 & 4$^{*}$ & 3{,}977 / 4 & 27{,}058 / 32{,}332 \\
 & Gemma & 200 & 193 & 2{,}631 / 770 & 4$^{*}$ & 7{,}036 / 4 & 29{,}125 / 32{,}955 \\
\addlinespace
GSM8K/FACTOR-News & Qwen & 200 & 195 & 2{,}339 / 733 & 196 & 6{,}201 / 2{,}090 & 25{,}533 / 28{,}488 \\
 & Llama & 200 & 195 & 2{,}447 / 745 & 196 & 3{,}632 / 1{,}180 & 23{,}729 / 27{,}605 \\
 & Gemma & 200 & 196 & 2{,}394 / 729 & 197 & 6{,}241 / 2{,}066 & 25{,}562 / 28{,}327 \\
\addlinespace
MATH/Wikitext & Qwen & 200 & 135 & 1{,}075 / 794 & 151 & 5{,}933 / 2{,}080 & 28{,}377 / 27{,}132 \\
 & Llama & 200 & 139 & 1{,}092 / 852 & 151 & 4{,}581 / 1{,}132 & 27{,}596 / 26{,}411 \\
 & Gemma & 200 & 137 & 1{,}076 / 813 & 151 & 5{,}905 / 2{,}132 & 27{,}549 / 26{,}935 \\
\addlinespace
DROP/Wikitext & Qwen & 168 & 137 & 1{,}839 / 845 & 139 & 5{,}070 / 2{,}183 & 30{,}388 / 29{,}844 \\
 & Llama & 185 & 149 & 2{,}106 / 957 & 150 & 3{,}355 / 1{,}267 & 32{,}421 / 31{,}417 \\
 & Gemma & 159 & 132 & 1{,}744 / 823 & 133 & 4{,}956 / 2{,}132 & 28{,}149 / 27{,}847 \\
\addlinespace
code/docstrings & Qwen & 200 & 40 & 2{,}747 / 115 & 37 & 439 / 549 & 17{,}929 / 17{,}135 \\
 & Llama & 200 & 41 & 2{,}747 / 121 & 38 & 348 / 419 & 18{,}038 / 17{,}366 \\
 & Gemma & 200 & 45 & 4{,}036 / 132 & 38 & 383 / 638 & 18{,}781 / 19{,}053 \\
\addlinespace
AGNews/Wikitext & Qwen & 82 & 22 & 67 / 129 & 29 & 263 / 354 & 4{,}178 / 4{,}990 \\
 & Llama & 75 & 19 & 62 / 120 & 26 & 127 / 168 & 3{,}787 / 4{,}464 \\
 & Gemma & 82 & 23 & 68 / 134 & 31 & 296 / 361 & 4{,}363 / 5{,}018 \\
\addlinespace
SQuAD/AGNews & Qwen & 118 & 13 & 86 / 46 & 24 & 373 / 175 & 4{,}170 / 4{,}496 \\
 & Llama & 119 & 8$^{*}$ & 59 / 57 & 15 & 133 / 128 & 4{,}384 / 4{,}620 \\
 & Gemma & 138 & 14 & 103 / 51 & 28 & 458 / 190 & 4{,}911 / 5{,}356 \\
\addlinespace
ARC/Wikitext & Qwen & 189 & 14 & 81 / 365 & 17 & 205 / 1{,}042 & 13{,}264 / 16{,}081 \\
 & Llama & 196 & 13 & 91 / 436 & 16 & 134 / 635 & 13{,}891 / 16{,}784 \\
 & Gemma & 190 & 11 & 91 / 362 & 15 & 220 / 1{,}043 & 13{,}768 / 16{,}053 \\
\addlinespace
StrategyQA/FACTOR-News & Qwen & 150 & 81 & 244 / 445 & 82 & 818 / 1{,}224 & 11{,}148 / 16{,}506 \\
 & Llama & 135 & 66 & 195 / 410 & 67 & 331 / 612 & 10{,}066 / 13{,}935 \\
 & Gemma & 149 & 77 & 240 / 445 & 78 & 819 / 1{,}223 & 10{,}958 / 15{,}884 \\
\addlinespace
TinyStories/AGNews & Qwen & 200 & 43 & 147 / 395 & 1$^{*}$ & 3 / 994 & 28{,}272 / 12{,}321 \\
 & Llama & 200 & 41 & 147 / 388 & 1$^{*}$ & 1 / 491 & 28{,}470 / 12{,}087 \\
 & Gemma & 200 & 124 & 661 / 398 & 1$^{*}$ & 4 / 1{,}018 & 28{,}287 / 12{,}480 \\
\bottomrule
\end{tabular}
\end{table}

\section{Risk stratification and held-out expansion}
\label{app:predictive}

\subsection{Prospective held-out domain-pair expansion}

The held-out analysis tests risk stratification: whether the screen assigns each
pair to the region it later occupies. We prospectively selected five of 11
candidate domain pairs
using tokenizer-side features from Qwen2.5-7B-Instruct,
Llama-3.2-3B-Instruct, and gemma-2-2b-it before examining any intervention
outcomes. The selection covers one high-risk prose pair, two low-risk prose
pairs, a structural boundary case, and an ambiguous case. The same CAI audit
was then applied to Qwen, Llama, and Gemma.

\begin{table}[H]
\centering
\small
\caption{Prospective selection of held-out pairs based on tokenizer-side features.
$\Delta_{\rm ws}$, $\Delta_{\rm chars}$, and target-type Jensen--Shannon
divergence (JS) are means over the three tokenizers.}
\label{tab:app_heldout_selection}
\resizebox{\linewidth}{!}{%
\begin{tabular}{lcrrrp{0.37\linewidth}}
\toprule
Pair & Pred. & $\Delta_{\rm ws}$ & $\Delta_{\rm chars}$ & JS & Selection reason \\
\midrule
GSM8K/TinyStories & high & +0.0739 & +0.8052 & 0.0952 & Large positive whitespace-rate and \mbox{chars/token} gaps in a prose pair \\
SQuAD/AGNews & low & +0.0077 & +0.3276 & 0.0074 & New QA/news pair with a small target-mixture gap \\
AGNews/Wikitext & low & \textminus0.0099 & +0.0055 & 0.0036 & Non-reasoning prose/prose baseline with small gaps \\
CodeSearchNet code/docs & boundary & +0.1414 & +0.0771 & 0.0537 & Structural whitespace flag with newline/indentation-heavy whitespace \\
TinyStories/AGNews & ambiguous & \textminus0.0202 & \textminus0.0383 & 0.0374 & Weak or inconsistent tokenizer-side signal \\
\bottomrule
\end{tabular}}
\end{table}

\begin{table}[H]
\centering
\scriptsize
\caption{Held-out audit outcomes across three model families. Each $n$ is the number of
matched sequences per side. Rows labelled boundary or ambiguous in advance are reported unscored. The
audit counts support within each corpus. Llama's SQuAD/AGNews $I$ uses an H=3
whitespace cell from 28 SQuAD and 38 AGNews sequences with a whitespace target,
and 8 matched pairs have one on both sides, below the paired rule of \tabref{tab:app_expansion_support}.}
\label{tab:app_heldout_outcomes}
\resizebox{\linewidth}{!}{%
\begin{tabular}{llcrrrr}
\toprule
Pair & Model & predicted region & $n$ & H=3 pooled & ws-excl. & audit $I$ [95\% CI] \\
\midrule
GSM8K/TinyStories & Qwen & high & 200 & +0.0254 & \textminus0.0291 & +0.6579 [0.6252, 0.6874] \\
 & Llama & high & 195 & +0.0407 & \textminus0.0280 & +0.7522 [0.7223, 0.7805] \\
 & Gemma & high & 200 & +0.0276 & \textminus0.0238 & +0.7088 [0.6844, 0.7360] \\
\addlinespace
SQuAD/AGNews & Qwen & low & 118 & +0.0245 & +0.0221 & +0.0573 [\textminus0.1302, 0.2287] \\
 & Llama & low & 119 & \textminus0.0017 & \textminus0.0120 & \textminus0.0727 [\textminus0.1763, 0.0225] \\
 & Gemma & low & 138 & +0.000019 & \textminus0.0082 & +0.0421 [\textminus0.0615, 0.1420] \\
\addlinespace
AGNews/Wikitext & Qwen & low & 82 & \textminus0.0192 & \textminus0.0174 & +0.0381 [\textminus0.1308, 0.2104] \\
 & Llama & low & 75 & \textminus0.0006 & +0.00003 & +0.1202 [\textminus0.0076, 0.2440] \\
 & Gemma & low & 82 & +0.0051 & +0.0050 & +0.1076 [0.0078, 0.2140] \\
\addlinespace
CodeSearchNet code/docs & Qwen & boundary & 200 & \textminus0.0051 & \textminus0.0194 & +0.0404 [\textminus0.0575, 0.1382] \\
 & Llama & boundary & 200 & \textminus0.0060 & \textminus0.0131 & +0.0481 [\textminus0.0244, 0.1325] \\
 & Gemma & boundary & 200 & +0.0166 & \textminus0.0352 & +0.1375 [0.0416, 0.2187] \\
\addlinespace
TinyStories/AGNews & Qwen & ambiguous & 200 & +0.0099 & +0.0205 & \textminus0.3583 [\textminus0.4357, \textminus0.2816] \\
 & Llama & ambiguous & 200 & +0.0060 & +0.0055 & \textminus0.1581 [\textminus0.2174, \textminus0.0916] \\
 & Gemma & ambiguous & 200 & +0.0060 & +0.0050 & \textminus0.0254 [\textminus0.0582, 0.0059] \\
\bottomrule
\end{tabular}}
\end{table}

All nine high/low rows fall within their predicted regions, and eight also meet
the corresponding confidence-interval prediction. Gemma on AGNews/Wikitext is the exception. Its interval lies strictly above zero, and its
point estimates, $I=0.108$ and a pooled CAI of $0.005$, meet the predefined low-risk
criteria. The boundary
results distinguish structural whitespace from prose-formatting whitespace. In the
ambiguous region, tokenizer-only features indicate that a full audit is needed.

All nine scored assignments are unchanged for any high
cutoff between $0.13$ and $0.65$ and any low cutoff on $|I|$ in the same range
paired with a pooled bound of $0.03$ or more, because every scored row labelled
low has $|I|$ below that range and every row labelled high has $|I|$ above it
(\tabref{tab:app_heldout_outcomes}).

\subsection{Reasoning-pair extension}
\label{app:reasoning_extension}

\begin{table}[H]
\centering
\scriptsize
\caption{MATH, DROP, and ARC extension against Wikitext across three model
families. Values are H=3 mean-blend top-1 CAI except $I$, which is the
interaction in Eq.~\ref{eq:audit}. Intervals use 2,000 sequence-level bootstrap
resamples. Across these nine rows and the three canonical rows of
\tabref{tab:canonical}, 12 model--pair rows, the whitespace-rate gap and pooled
CAI have Spearman $\rho=0.9371$, with agreement in 60 of 66 pairwise
orderings.}
\label{tab:app_reasoning_extension}
\resizebox{\linewidth}{!}{%
\begin{tabular}{llrrrrrrr}
\toprule
Pair & Model & $n$/side & pooled & ws & ws-excl. & ref-rw. & $I$ & 95\% CI \\
\midrule
MATH/Wikitext & Qwen & 200 & \textminus0.0087 & +0.4324 & \textminus0.0242 & \textminus0.0058 & +0.5183 & [0.4504, 0.5841] \\
 & Llama & 200 & +0.0062 & +0.6431 & \textminus0.0314 & +0.0027 & +0.6895 & [0.6249, 0.7527] \\
 & Gemma & 200 & +0.0120 & +0.5188 & \textminus0.0079 & +0.0124 & +0.5878 & [0.5285, 0.6471] \\
\addlinespace
DROP/Wikitext & Qwen & 168 & +0.0078 & +0.4327 & \textminus0.0211 & \textminus0.0105 & +0.4265 & [0.3723, 0.4794] \\
 & Llama & 185 & +0.0179 & +0.7584 & \textminus0.0283 & \textminus0.0079 & +0.7605 & [0.7217, 0.7985] \\
 & Gemma & 159 & +0.0123 & +0.4588 & \textminus0.0143 & \textminus0.0019 & +0.4925 & [0.4568, 0.5277] \\
\addlinespace
ARC/Wikitext & Qwen & 189 & +0.0038 & \textminus0.0205 & +0.0051 & +0.0044 & \textminus0.0611 & [\textminus0.1432, 0.0341] \\
 & Llama & 196 & \textminus0.0089 & +0.0159 & \textminus0.0241 & \textminus0.0145 & +0.0377 & [\textminus0.0279, 0.0938] \\
 & Gemma & 190 & +0.0074 & \textminus0.0281 & +0.0070 & +0.0051 & \textminus0.0004 & [\textminus0.0646, 0.0659] \\
\bottomrule
\end{tabular}}
\end{table}

MATH and DROP yield six positive interactions whose intervals exclude zero.
Their pooled estimates are positive in five of six rows, all except Qwen/MATH at
$-0.0087$. ARC is the low-risk control: all three of its
interaction intervals include zero, and its whitespace cell is positive in one
of the three models and negative in the other two. The screen separates the
two medium-risk pairs from the low-risk control. \tabref{tab:app_screen_inputs}
gives its inputs as means over the three tokenizers. Under
\tabref{tab:screen_rule}, MATH and DROP miss the high rule on $\DeltaWS$ and
meet the medium rule, and ARC meets the low rule.

\begin{table}[H]
\centering
\small
\caption{Screen inputs of the reasoning and DoLa-motivated pairs, as means over
the three tokenizers, and the label each pair receives under
\tabref{tab:screen_rule}.}
\label{tab:app_screen_inputs}
\begin{tabular}{lrrrl}
\toprule
Pair & $\DeltaWS$ & $\DeltaChars$ & JS & Label \\
\midrule
MATH/Wikitext & +0.0080 & +1.8287 & 0.0252 & medium \\
DROP/Wikitext & +0.0301 & +0.5212 & 0.0126 & medium \\
ARC/Wikitext & \textminus0.0167 & +0.0923 & 0.0127 & low \\
GSM8K/FACTOR-News & +0.0619 & +1.1306 & 0.0372 & high \\
StrategyQA/FACTOR-News & \textminus0.0068 & +0.0865 & 0.0040 & low \\
\bottomrule
\end{tabular}
\end{table}

Across the canonical, MATH, DROP, and ARC pairs (12 model--pair rows), the
tokenizer-only whitespace-rate gap and pooled H=3 CAI show strong rank
agreement. Two of the three families, Llama and Gemma, reproduce the exact
four-pair order, and Qwen swaps only MATH and ARC, as
\tabref{tab:app_reasoning_extension} shows.

\subsection{DoLa-motivated corpus-pair audit}
\label{app:dola_update}

We apply CAI to two corpus pairs built from datasets studied by DoLa:
GSM8K/FACTOR-News and StrategyQA/FACTOR-News. Both pairs were classified in
advance using tokenizer-only criteria. On the tokenizer means in
\tabref{tab:app_screen_inputs}, GSM8K/FACTOR-News meets the high rule of
\tabref{tab:screen_rule}, and StrategyQA/FACTOR-News meets the low rule.
FACTOR-News sequences are drawn from the 1,036-row test split of a
public NEWS-FACTOR mirror. The audit compares these corpora using
our interventions. It does not evaluate DoLa's layer selection
or decoding gains.

\begin{table}[H]
\centering
\scriptsize
\caption{DoLa-motivated corpus audit. The first six rows are prospective. The
last three are GSM8K/Wikitext calibration replications and are excluded from the
41-condition coverage count. Intervals use 10,000 sequence-level bootstrap resamples.}
\label{tab:app_dola}
\resizebox{\linewidth}{!}{%
\begin{tabular}{llrrrrrrr}
\toprule
Pair & Model & $n$/side & pooled & ws & ws-excl. & ref-rw. & $I$ & 95\% CI \\
\midrule
GSM8K/FACTOR-News & Qwen & 200 & +0.0340 & +0.5374 & \textminus0.0182 & \textminus0.0052 & +0.5289 & [0.4832, 0.5733] \\
 & Llama & 200 & +0.0467 & +0.7878 & \textminus0.0266 & \textminus0.0060 & +0.7883 & [0.7494, 0.8270] \\
 & Gemma & 200 & +0.0351 & +0.7598 & \textminus0.0207 & \textminus0.0043 & +0.7566 & [0.7196, 0.7932] \\
\addlinespace
StrategyQA/FACTOR-News & Qwen & 150 & \textminus0.0048 & \textminus0.0141 & \textminus0.0055 & \textminus0.0080 & \textminus0.0507 & [\textminus0.1023, 0.0034] \\
 & Llama & 135 & \textminus0.0070 & \textminus0.0118 & \textminus0.0075 & \textminus0.0070 & \textminus0.0409 & [\textminus0.0982, 0.0119] \\
 & Gemma & 149 & +0.0009 & +0.0335 & +0.0011 & +0.0017 & +0.0337 & [\textminus0.0078, 0.0738] \\
\midrule
GSM8K/Wikitext calibration & Qwen & 200 & +0.0260 & +0.4525 & \textminus0.0216 & \textminus0.0114 & +0.4532 & [0.3943, 0.5097] \\
 & Llama & 200 & +0.0416 & +0.8137 & \textminus0.0300 & \textminus0.0106 & +0.8169 & [0.7781, 0.8554] \\
 & Gemma & 200 & +0.0432 & +0.7135 & \textminus0.0174 & +0.0072 & +0.7117 & [0.6737, 0.7494] \\
\bottomrule
\end{tabular}}
\end{table}

All six prospective outcomes fall within their predicted regions. The three
GSM8K/FACTOR rows show the high-risk pattern, including negative values under
both mixture corrections, and all three StrategyQA/FACTOR interaction
intervals contain zero. The result flags a target-mixture risk for
GSM8K/FACTOR-News.
\begin{table}[H]
\centering
\small
\caption{Aggregate sign reversals among rows with positive pooled estimates.
The two right-hand columns count reversals under whitespace removal
(``ws-excl.'')\@ and under the reference mixture (``ref.-rw.'').
The high- and medium-risk set comprises the canonical,
GSM8K/TinyStories, pooled-positive MATH/DROP, and GSM8K/FACTOR-News rows.}
\label{tab:app_reversals}
\begin{tabular}{lrrr}
\toprule
Set & pooled-positive rows & ws-excl. & ref.-rw. \\
\midrule
High/medium risk & 14 & 14 & 11 \\
Other decomposed rows & 10 & 2 & n/a \\
\bottomrule
\end{tabular}

\vspace{2pt}
\begin{minipage}{0.94\linewidth}
\footnotesize
In the first row, whitespace removal reverses the sign of all 14 estimates.
Reference reweighting reverses 11 of 14. Its exceptions are Llama/MATH,
Gemma/MATH, and Gemma on the canonical pair. Outside that set, 2 of 10 pooled-positive rows reverse under whitespace removal:
Gemma on CodeSearchNet code/docs and Gemma on SQuAD/AGNews. The latter has an
uncorrected estimate of $+1.9\times10^{-5}$.
\end{minipage}
\end{table}

The reversal counts cover rows whose pooled estimate starts positive. Among the
9 rows whose pooled estimate starts negative, 8 stay negative once whitespace is
excluded, and Llama on AGNews/Wikitext moves just above zero
(\tabref{tab:app_expansion_intervals}).

\subsection{Calibration-screen rows by model and domain pair}

\begin{table}[H]
\centering
\small
\caption{Calibration-screen risk rows. $\Delta_{\rm chars}$ is the mean number of
characters per token in $\Ndom$ minus $\Rdom$. $\Delta_{\rm ws}$ is the target
whitespace rate in $\Rdom$ minus $\Ndom$.}
\label{tab:app_predictive_rows}
\resizebox{\linewidth}{!}{%
\begin{tabular}{llrrrr}
\toprule
Domain pair & Model & $\Delta_{\rm chars}$ & $\Delta_{\rm ws}$ & audit $I$ & $p$ \\
\midrule
GSM8K/Wikitext & Qwen2.5-7B & 1.1960 & 0.0559 & 0.5193 & $<10^{-4}$ \\
GSM8K/Wikitext & Llama-3.2-3B & 0.8890 & 0.0664 & 0.8133 & $<10^{-4}$ \\
GSM8K/Wikitext & gemma-2-2b & 1.3120 & 0.0551 & 0.6850 & $<10^{-4}$ \\
GSM8K/Wikitext & gemma-2-9b & 1.2610 & 0.0527 & 0.6034 & $<10^{-4}$ \\
GSM8K/Wikitext & Hermes-3-Llama-3.1-8B & 0.8292 & 0.0642 & 0.7968 & $<10^{-4}$ \\
Code/Wikitext & Qwen2.5-7B & \textminus0.0595 & 0.1043 & \textminus0.0272 & 1.0 \\
Code/Wikitext & Llama-3.2-3B & \textminus0.0121 & 0.1026 & 0.0684 & $<0.001$ \\
Code/Wikitext & gemma-2-2b & 0.9211 & 0.1505 & 0.2846 & $<0.001$ \\
HotpotQA/AGNews & Qwen2.5-7B & 0.4339 & 0.0134 & \textminus0.1909 & 1.0 \\
HotpotQA/AGNews & Llama-3.2-3B & 0.2201 & 0.0111 & \textminus0.0079 & 1.0 \\
HotpotQA/AGNews & gemma-2-2b & 0.4926 & 0.0392 & 0.1464 & $<0.001$ \\
\bottomrule
\end{tabular}}
\end{table}

\begin{figure}[H]
\centering
\includegraphics[width=\linewidth]{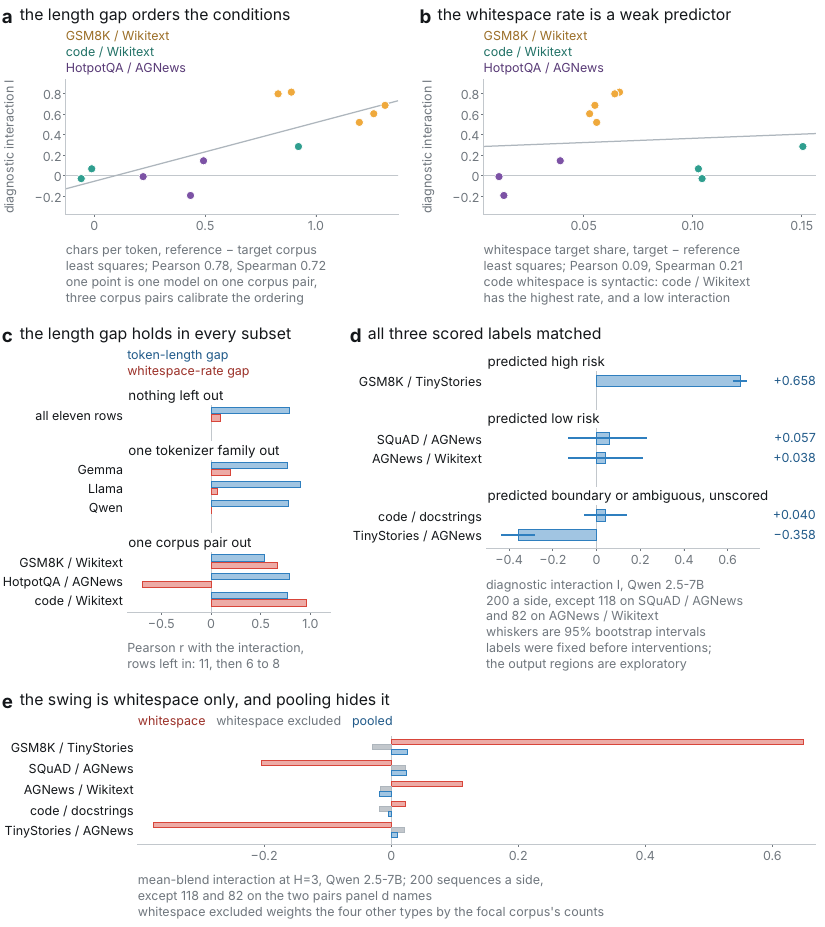}
\caption{The full screen figure. Panels \textbf{a} and \textbf{b} plot the
diagnostic interaction $I$ for the eleven calibration rows against the
characters-per-token gap and against the target whitespace-rate gap. Panel
\textbf{c} gives both Pearson correlations over all eleven rows and after leaving
out one tokenizer family or one corpus pair. Panel \textbf{d} shows $I$ with its
95\% bootstrap interval for the five held-out pairs on Qwen2.5-7B, grouped by
label, and panel \textbf{e} compares three H=3 mean-blend quantities for each
held-out pair: the whitespace cell, the whitespace-excluded summary, and the
pooled value. \figref{fig:screen} repeats panels \textbf{a} to \textbf{c}, and
its panel \textbf{d} extends \textbf{d} to all three primary models and the two
DoLa pairs.}
\label{fig:screen_full}
\end{figure}

\figref{fig:screen_full}a and \figref{fig:screen_full}b plot all eleven calibration-screen
rows against two tokenizer-only predictors: characters per token in $\Ndom$
minus $\Rdom$, and the target whitespace rate in $\Rdom$ minus $\Ndom$.
\tabref{tab:app_predictive_rows} lists the rows. Every GSM8K/Wikitext row has
$p<10^{-4}$, and on the other two pairs the positive rows have $p<0.001$ and the
null or reversed rows have $p=1.0$.
\tabref{tab:app_domain_pair_summary} makes the clustering explicit by
aggregating the 11 model rows into three domain-pair regions. These summaries
show where the risk pattern concentrates.

\begin{table}[H]
\centering
\small
\caption{Domain-pair-level aggregation for the calibration screen. Each cell
reports mean [min, max] over model rows in that domain pair.}
\label{tab:app_domain_pair_summary}
\resizebox{\linewidth}{!}{%
\begin{tabular}{lrrrr}
\toprule
Domain pair & $n$ & $\Delta_{\rm chars}$ & $\Delta_{\rm ws}$ & audit $I$ \\
\midrule
GSM8K/Wikitext & 5 & 1.097 [0.829, 1.312] & 0.0589 [0.0527, 0.0664] & 0.684 [0.519, 0.813] \\
Code/Wikitext & 3 & 0.283 [\textminus0.060, 0.921] & 0.1191 [0.1026, 0.1505] & 0.109 [\textminus0.027, 0.285] \\
HotpotQA/AGNews & 3 & 0.382 [0.220, 0.493] & 0.0212 [0.0111, 0.0392] & \textminus0.018 [\textminus0.191, 0.146] \\
\bottomrule
\end{tabular}}
\end{table}

The rows form three risk regions. GSM8K/Wikitext is a high-risk prose pair,
where both predictors agree and the audit interaction is large in every model.
HotpotQA/AGNews is the lower-risk prose pair, with smaller whitespace-rate and
characters-per-token gaps, yielding a null or reversed audit interaction in two
rows and a weakly positive interaction for Gemma. Code/Wikitext forms a boundary region: its target whitespace rate is high, its
whitespace is structural, and its audit magnitudes are mixed.

\begin{table}[H]
\centering
\small
\caption{Calibration-screen risk regions. Held-out domain-pair outcomes appear in
\tabref{tab:app_heldout_outcomes}.}
\label{tab:risk_regions}
\begin{tabular}{p{0.24\linewidth}p{0.26\linewidth}p{0.39\linewidth}}
\toprule
Region & Observed predictor range & Interpretation \\
\midrule
GSM8K/Wikitext & $\Delta_{\rm ws}=0.053$--$0.066$,\newline $\Delta_{\rm chars}=0.83$--$1.31$,\newline $I=0.52$--$0.81$ & High-risk prose pair. Both predictors agree, and all five rows have zero non-positive bootstrap draws \\
Code/Wikitext & $\Delta_{\rm ws}=0.103$--$0.150$,\newline $\Delta_{\rm chars}=-0.06$--$0.92$,\newline $I=-0.03$--$0.28$ & Boundary case. Whitespace targets are frequent and mostly newlines or indentation \\
HotpotQA/AGNews & $\Delta_{\rm ws}=0.011$--$0.039$,\newline $\Delta_{\rm chars}=0.22$--$0.49$,\newline $I=-0.19$--$0.15$ & Lower-risk prose pair. Two rows are null or reversed, and one Gemma row is weakly positive \\
\bottomrule
\end{tabular}
\end{table}

\subsection{Correlation and leave-one sensitivity}

\figref{fig:screen_full}c compares the all-row correlations with values recomputed
after leaving out one tokenizer family or one corpus pair.
The characters-per-token correlation stays positive in every leave-one-out subset.
The target whitespace rate tracks risk in prose but saturates on code, where
whitespace is structural.

\begin{table}[H]
\centering
\small
\caption{Predictor correlations over the 11 model rows. The rows form three domain-pair clusters.}
\label{tab:app_predictive_corr}
\begin{tabular}{lrr}
\toprule
Predictor & Pearson $r$ & Spearman $\rho$ \\
\midrule
Target whitespace-rate gap & 0.093 & 0.209 \\
Chars-per-token gap & 0.783 & 0.718 \\
\bottomrule
\end{tabular}
\end{table}

\begin{table}[H]
\centering
\small
\caption{Leave-one-model-lineage sensitivity. The characters-per-token
correlations stay positive with any one lineage left out.}
\label{tab:app_leave_family}
\begin{tabular}{lrrrrr}
\toprule
Dropped lineage & $n$ & Pearson ws & Spearman ws & Pearson chars & Spearman chars \\
\midrule
Gemma & 7 & 0.192 & 0.179 & 0.771 & 0.643 \\
Llama & 7 & 0.064 & 0.214 & 0.901 & 0.964 \\
Qwen & 8 & 0.006 & 0.310 & 0.778 & 0.619 \\
\bottomrule
\end{tabular}
\end{table}

\begin{table}[H]
\centering
\small
\caption{Leave-one-domain-pair sensitivity. Removing the code pair restores the
direct relationship with target whitespace rate because whitespace serves a
different function in code and prose.}
\label{tab:app_leave_pair}
\begin{tabular}{lrrrrr}
\toprule
Dropped pair & $n$ & Pearson ws & Spearman ws & Pearson chars & Spearman chars \\
\midrule
GSM8K/Wikitext & 6 & 0.666 & 0.486 & 0.538 & 0.600 \\
HotpotQA/AGNews & 8 & \textminus0.689 & \textminus0.524 & 0.784 & 0.429 \\
Code/Wikitext & 8 & 0.961 & 0.905 & 0.771 & 0.619 \\
\bottomrule
\end{tabular}
\end{table}

These sensitivity results support the two-predictor screen. The target
whitespace rate directly tracks risk for prose pairs, and characters per token
provides a more reliable prospective proxy for code-like text and other corpora
in which whitespace has domain-specific structure.

\subsection{Code boundary details}

The code boundary shows that whitespace frequency and whitespace function are
separate quantities. Code has a high target whitespace rate, and the
tokenizer-only subtype analysis identifies those targets as mostly newlines or
indentation. The corresponding GSM8K/Wikitext whitespace targets are plain
spaces, and the held-out CodeSearchNet code/docs pair shows the same
qualitative boundary within one source family.

\paragraph{Code boundary vignette.}
The bracketed target in this code excerpt is a structural line break:

{\ttfamily\footnotesize .X = vec1.X * vec2.X\quad new\_vec.Y = vec1.Y *
vec2.Y[[\textbackslash n]]\quad return new\_vec}

This target is structural whitespace, which the subtype analysis in
\tabref{tab:app_ws_subtypes} separates from the plain spaces of prose.

\subsection{Tokenizer-only extension details}

\begin{table}[H]
\centering
\small
\caption{Target rates for Code/Wikitext.}
\label{tab:app_code_rates}
\resizebox{\linewidth}{!}{%
\begin{tabular}{lrrrrrrr}
\toprule
Model & $R$ ws rate & $N$ ws rate & $R$ targets & $N$ targets & $R$ chars/token & $N$ chars/token & $\Delta_{\rm chars}$ \\
\midrule
Qwen2.5-7B & 0.1319 & 0.0277 & 23307 & 26118 & 4.4728 & 4.4132 & \textminus0.0595 \\
Llama-3.2-3B & 0.1332 & 0.0306 & 23297 & 25991 & 4.4909 & 4.4788 & \textminus0.0121 \\
gemma-2-2b & 0.1786 & 0.0281 & 25321 & 26528 & 3.6877 & 4.6089 & 0.9211 \\
\bottomrule
\end{tabular}}
\end{table}

\begin{table}[H]
\centering
\small
\caption{Target rates for HotpotQA/AGNews.}
\label{tab:app_hotpot_rates}
\resizebox{\linewidth}{!}{%
\begin{tabular}{lrrrrrrrr}
\toprule
Model & $n_R/n_N$ & $R$ ws rate & $N$ ws rate & $R$ targets & $N$ targets & $R$ chars/token & $N$ chars/token & $\Delta_{\rm chars}$ \\
\midrule
Qwen2.5-7B & 146/146 & 0.0301 & 0.0167 & 5789 & 6403 & 4.2404 & 4.6743 & 0.4339 \\
Llama-3.2-3B & 135/135 & 0.0292 & 0.0181 & 5336 & 5856 & 4.4513 & 4.6715 & 0.2201 \\
gemma-2-2b & 161/161 & 0.0565 & 0.0173 & 6179 & 6950 & 4.2427 & 4.7354 & 0.4926 \\
\bottomrule
\end{tabular}}
\end{table}

\begin{table}[H]
\centering
\small
\caption{Tokenizer statistics for the additional GSM8K/Wikitext
audit models.}
\label{tab:app_gsm_cpu_rates}
\resizebox{\linewidth}{!}{%
\begin{tabular}{lrrrrrrr}
\toprule
Model & $R$ ws rate & $N$ ws rate & $R$ targets & $N$ targets & $R$ chars/token & $N$ chars/token & $\Delta_{\rm chars}$ \\
\midrule
gemma-2-9b-it & 0.0813 & 0.0286 & 26947 & 28403 & 3.3460 & 4.6070 & 1.2610 \\
Hermes-3-Llama-3.1-8B & 0.0945 & 0.0303 & 24743 & 26392 & 3.6596 & 4.4888 & 0.8292 \\
\bottomrule
\end{tabular}}
\end{table}

\subsection{Whitespace subtype and classifier robustness}
\label{app:classifier_robustness}

The main classifier uses a coarse category boundary. A decoded token is whitespace only
when its entire decoded text is whitespace. Otherwise, the text is stripped
before the classifier assigns a number, punctuation, function-word, or
content-word label. Leading-space BPE or SentencePiece tokens therefore retain
their non-whitespace category after stripping.

\begin{table}[H]
\centering
\small
\caption{Whitespace subtype robustness at H=3, averaged over the Qwen2.5-7B,
Llama-3.2-3B, and gemma-2-2b tokenizers. ``ws rate'' is the all-whitespace
target rate. Subtype columns are proportions within whitespace targets. ``lead
cont.'' is the proportion of content-word targets whose decoded token begins
with whitespace and is classified after stripping. These tokenizer-only rates use their own 200 length-matched sequences per corpus,
so they differ slightly from the canonical sample's in \tabref{tab:app_weights} and the screen's
in \tabref{tab:app_code_rates}.}
\label{tab:app_ws_subtypes}
\resizebox{\linewidth}{!}{%
\begin{tabular}{llrrrrrr}
\toprule
Pair & Domain & ws rate & plain & newline & multi/indent & other & lead cont. \\
\midrule
GSM8K/Wikitext & GSM8K & 0.0888 & 1.000 & 0.000 & 0.000 & 0.000 & 0.9086 \\
GSM8K/Wikitext & Wikitext & 0.0276 & 1.000 & 0.000 & 0.000 & 0.000 & 0.8598 \\
Code/Wikitext & Code & 0.1454 & 0.039 & 0.266 & 0.687 & 0.007 & 0.4644 \\
Code/Wikitext & Wikitext & 0.0281 & 1.000 & 0.000 & 0.000 & 0.000 & 0.8647 \\
\bottomrule
\end{tabular}}
\end{table}

The subtype analysis clarifies the code boundary because whitespace targets in
the prose rows are plain-space tokens, whereas the additional code whitespace
consists mainly of newlines and indentation. Whitespace function enters the
screen alongside its frequency. The intervention analysis uses
the coarser split between all-whitespace and non-whitespace targets. Under the
canonical map every non-whitespace subclass is non-positive for the H=3
mean-blend artifact, and the one positive cell under the other maps is Gemma's
content cell in Appendix~\ref{app:classifier_policy}.

\subsection{Vocabulary-policy intervention audit}
\label{app:classifier_policy}

The classifier-policy analysis keeps the sequences, activations, operator,
horizon, and bootstrap procedure fixed while changing the vocabulary map. The
canonical policy is the five-way map used throughout the paper.
The lead-only policy keeps in the content class the content words whose decoded
string begins with whitespace, and the nolead policy keeps the others. Each policy moves its other content words into the function class, so
in their rows of \tabref{tab:app_classifier_cells} the function column pools
the function words with the moved content words. The
adversarial policy assigns every leading-whitespace token to the whitespace
class, which absorbs most leading-space words.

\begin{table}[H]
\centering
\scriptsize
\caption{H=3 mean-blend top-1 CAI under four fixed classifier policies, using
200 sequences per side and 10,000 resamples. ``n/a'' denotes an empty or undefined
function-word cell under the adversarial map. Under that map, whitespace CAI
shrinks by factors of 8.2, 8.7, and 8.4 for Qwen, Llama, and Gemma. The shared
positive Gemma nolead/adversarial content cell is $+0.003$, with intervals
$[-0.001,0.010]$ and $[-0.0007,0.010]$.}
\label{tab:app_classifier_cells}
\resizebox{\linewidth}{!}{%
\begin{tabular}{llrrrrr}
\toprule
Model & policy & whitespace & content & number & punctuation & function \\
\midrule
Qwen & canonical & +0.4525 & \textminus0.0084 & \textminus0.0107 & \textminus0.1026 & \textminus0.0216 \\
 & lead-only & +0.4525 & \textminus0.0090 & \textminus0.0107 & \textminus0.1026 & \textminus0.0182 \\
 & nolead & +0.4525 & \textminus0.0018 & \textminus0.0107 & \textminus0.1026 & \textminus0.0137 \\
 & adversarial & +0.0552 & \textminus0.0018 & \textminus0.0107 & \textminus0.0624 & n/a \\
\addlinespace
Llama & canonical & +0.8137 & \textminus0.0081 & \textminus0.0058 & \textminus0.1973 & \textminus0.0140 \\
 & lead-only & +0.8137 & \textminus0.0082 & \textminus0.0058 & \textminus0.1973 & \textminus0.0133 \\
 & nolead & +0.8137 & \textminus0.0059 & \textminus0.0058 & \textminus0.1973 & \textminus0.0106 \\
 & adversarial & +0.0940 & \textminus0.0059 & \textminus0.0058 & \textminus0.2682 & n/a \\
\addlinespace
Gemma & canonical & +0.7135 & \textminus0.0023 & \textminus0.0319 & \textminus0.0548 & \textminus0.0109 \\
 & lead-only & +0.7135 & \textminus0.0025 & \textminus0.0319 & \textminus0.0548 & \textminus0.0096 \\
 & nolead & +0.7135 & +0.0030 & \textminus0.0319 & \textminus0.0548 & \textminus0.0056 \\
 & adversarial & +0.0852 & +0.0030 & \textminus0.0319 & \textminus0.0541 & n/a \\
\bottomrule
\end{tabular}}
\par\smallskip
\begin{minipage}{0.94\linewidth}
\scriptsize\emph{H=1 nolead content.} The corresponding rows are Qwen
$+0.077$ [$0.041,0.115$] ($n=174$), Llama $+0.116$ [$0.070,0.160$]
($n=191$), and Gemma $+0.068$ [$0.041,0.095$] ($n=184$).
\end{minipage}
\end{table}

\begin{table}[H]
\centering
\scriptsize
\caption{Vocabulary class sizes for Table~\ref{tab:app_classifier_cells}.
Each entry counts vocabulary types. Under the
adversarial map, whitespace accounts for approximately 36\%, 46\%, and 49\% in
Qwen, Llama, and Gemma.}
\label{tab:app_classifier_sizes}
\resizebox{\linewidth}{!}{%
\begin{tabular}{llrrrrr}
\toprule
Model & policy & whitespace & content & number & punctuation & function \\
\midrule
Qwen & canonical & 441 & 146438 & 28 & 4106 & 630 \\
 & lead-only & 441 & 52484 & 28 & 4106 & 94584 \\
 & nolead & 441 & 93954 & 28 & 4106 & 53114 \\
 & adversarial & 54409 & 93954 & 28 & 2985 & 267 \\
\addlinespace
Llama & canonical & 530 & 121495 & 1206 & 4136 & 633 \\
 & lead-only & 530 & 57295 & 1206 & 4136 & 64833 \\
 & nolead & 530 & 64200 & 1206 & 4136 & 57928 \\
 & adversarial & 59289 & 64200 & 1206 & 3038 & 267 \\
\addlinespace
Gemma & canonical & 152 & 251173 & 397 & 3658 & 620 \\
 & lead-only & 152 & 124041 & 397 & 3658 & 127752 \\
 & nolead & 152 & 127132 & 397 & 3658 & 124661 \\
 & adversarial & 125407 & 127132 & 350 & 2805 & 306 \\
\bottomrule
\end{tabular}}
\end{table}

Under the adversarial map, the whitespace vocabulary class expands sharply and
its CAI shrinks, yet whitespace remains the largest positive cell in all three
models and the only positive cell in Qwen and Llama. In some rows the punctuation cell is the most negative cell and has the largest
absolute value. The four maps define 45 non-whitespace cells, and a cell that a policy
leaves unchanged repeats its estimate, which leaves 27 distinct estimates. One
of them is positive, Gemma's content estimate under the nolead and adversarial
maps, and both of its intervals include zero
(\tabref{tab:app_classifier_cells}).

At H=1 under mean-blend, the nolead content cell is positive with an interval
excluding zero in all three models, as the note in
\tabref{tab:app_classifier_cells} reports.

\clearpage
\section{The one-token-ahead content-token cell and independent corroboration}
\label{app:residual}

\subsection{Canonical H=1 content-token effect-size rows}

\begin{table}[H]
\centering
\small
\caption{Scale diagnostics for the four canonical H=1 content-token
mild-resample rows. CAI is the absolute top-1 interaction. Rel.\% is
$100\times$ CAI divided by the reasoning-domain control top-1 level.}
\label{tab:app_effect_size_raw}
\resizebox{\linewidth}{!}{%
\begin{tabular}{lrrrrrrrrrr}
\toprule
Model & CAI & $d$ & rel.\% & ctrl $R$ & ctrl $N$ & early $R$ & early $N$ & late $R$ & late $N$ & $R/N$ chars \\
\midrule
Qwen2.5-7B & 0.0107 & 0.3507 & 1.22 & 0.8768 & 0.3904 & 0.0499 & 0.0350 & 0.0735 & 0.0479 & 3.260/4.456 \\
Llama-3.2-3B & 0.0117 & 0.3898 & 1.89 & 0.6194 & 0.3658 & 0.0604 & 0.0398 & 0.0858 & 0.0535 & 3.656/4.545 \\
gemma-2-2b & 0.0091 & 0.3407 & 1.51 & 0.6004 & 0.3277 & 0.0480 & 0.0289 & 0.0673 & 0.0392 & 3.335/4.647 \\
Hermes-3-Llama-3.1-8B & 0.0103 & 0.3370 & 1.64 & 0.6297 & 0.4262 & 0.0658 & 0.0490 & 0.0909 & 0.0638 & 3.656/4.545 \\
\bottomrule
\end{tabular}}
\end{table}

These rows give the scale of the H=1 content cell. The relative percentage
divides the interaction by the GSM8K control accuracy, the higher of the two
control levels in every row, and Appendix~\ref{app:h1_control_split} traces the
cell to that difference in control accuracy.

\subsection{Five-instance H=1 content-word results}

\begin{table}[H]
\centering
\small
\caption{Supporting H=1 content-token mild-resample results under the same
protocol. ``Positive
cell?'' means that the reported 95\% bootstrap interval is strictly positive.
The intervals use 1,000 sequence-bootstrap resamples. Table~\ref{tab:app_effect_size_raw}
gives the four-row Cohen's $d$ result with its paired per-sequence diagnostics.}
\label{tab:app_residual_support}
\begin{tabular}{lrrrr}
\toprule
Model & CAI & 95\% CI & $n_R/n_N$ & Positive cell? \\
\midrule
Qwen2.5-7B-Instruct & 0.0103 & [0.0050, 0.0149] & 250/250 & yes \\
Llama-3.2-3B-Instruct & 0.0142 & [0.0089, 0.0193] & 250/250 & yes \\
gemma-2-2b-it & 0.0086 & [0.0044, 0.0135] & 250/250 & yes \\
OLMo-2-1124-7B-Instruct & 0.0148 & [0.0092, 0.0207] & 250/250 & yes \\
Qwen3-1.7B & 0.0125 & [0.0076, 0.0172] & 250/250 & yes \\
\bottomrule
\end{tabular}
\end{table}

\subsection{The other four target types in the same H=1 series}

\tabref{tab:app_h1_types} conditions the same five-instance H=1 mild-resample
series on the four remaining target types, so the one-token-ahead cells can be
read type by type. With a positive cell defined as a strictly positive 95\% bootstrap interval,
number targets give four positive
cells, punctuation gives three, function words give one, and whitespace gives
none. Under this operator and horizon every positive cell sits on a
non-whitespace type, which inverts the H=3 mean-blend ordering.

\begin{table}[H]
\centering
\small
\caption{H=1 mild-resample top-1 CAI on GSM8K/Wikitext for the four target
types outside \tabref{tab:app_residual_support}, under the same protocol and the
same 250 sequences per corpus. $n_\Rdom/n_\Ndom$ counts the sequences supplying a
target of the row's type, so whitespace and number rows draw on fewer Wikitext
sequences. ``Positive cell?''\@ means that the reported 95\% bootstrap interval, from 1,000
sequence-bootstrap resamples, is strictly positive.}
\label{tab:app_h1_types}
\begin{tabular}{llrrrr}
\toprule
Target type & Model & CAI & 95\% CI & $n_\Rdom/n_\Ndom$ & Positive cell? \\
\midrule
whitespace & Qwen2.5-7B-Instruct & \textminus0.0235 & [\textminus0.0465, \textminus0.0030] & 250/181 & no \\
whitespace & Llama-3.2-3B-Instruct & \textminus0.0028 & [\textminus0.0208, 0.0155] & 250/179 & no \\
whitespace & gemma-2-2b-it & \textminus0.0076 & [\textminus0.0208, 0.0046] & 250/182 & no \\
whitespace & OLMo-2-1124-7B-Instruct & \textminus0.0073 & [\textminus0.0303, 0.0107] & 250/179 & no \\
whitespace & Qwen3-1.7B & 0.0038 & [\textminus0.0081, 0.0166] & 250/181 & no \\
\midrule
number & Qwen2.5-7B-Instruct & 0.0038 & [\textminus0.0054, 0.0134] & 250/182 & no \\
number & Llama-3.2-3B-Instruct & 0.0629 & [0.0465, 0.0816] & 250/180 & yes \\
number & gemma-2-2b-it & 0.0445 & [0.0352, 0.0551] & 250/184 & yes \\
number & OLMo-2-1124-7B-Instruct & 0.0462 & [0.0268, 0.0634] & 250/180 & yes \\
number & Qwen3-1.7B & 0.0293 & [0.0172, 0.0416] & 250/182 & yes \\
\midrule
punctuation & Qwen2.5-7B-Instruct & 0.0007 & [\textminus0.0088, 0.0109] & 250/250 & no \\
punctuation & Llama-3.2-3B-Instruct & 0.0073 & [\textminus0.0013, 0.0162] & 250/250 & no \\
punctuation & gemma-2-2b-it & 0.0234 & [0.0160, 0.0309] & 250/250 & yes \\
punctuation & OLMo-2-1124-7B-Instruct & 0.0183 & [0.0071, 0.0287] & 250/250 & yes \\
punctuation & Qwen3-1.7B & 0.0253 & [0.0159, 0.0350] & 250/250 & yes \\
\midrule
function word & Qwen2.5-7B-Instruct & 0.0039 & [\textminus0.0017, 0.0095] & 250/250 & no \\
function word & Llama-3.2-3B-Instruct & 0.0101 & [0.0044, 0.0157] & 250/250 & yes \\
function word & gemma-2-2b-it & \textminus0.0004 & [\textminus0.0047, 0.0037] & 250/250 & no \\
function word & OLMo-2-1124-7B-Instruct & 0.0025 & [\textminus0.0040, 0.0091] & 250/250 & no \\
function word & Qwen3-1.7B & \textminus0.0030 & [\textminus0.0073, 0.0012] & 250/250 & no \\
\bottomrule
\end{tabular}
\end{table}

\subsection{The content cell split by control-pass correctness}
\label{app:h1_control_split}

The split needs each target's token ID and its top-1 correctness under the
control, early-band, and late-band passes. A second evaluation pass of the H=1
series with the same code path, data, length matching, donor derangement, bands,
and $\alpha=0.2$ records them, and it reproduces every H=1 top-1 type cell of the
series exactly. Its
intervals come from 2,000 sequence-bootstrap draws that recompute every stratum
from the same draw, so they differ slightly from the 1,000-draw intervals of
\tabref{tab:app_residual_support} and \tabref{tab:app_h1_types}.

For a target type, let $p_C$ be the share of its targets in corpus $C$ that the
control pass predicts at top-1. The operator acts on those correct predictions:
it removes 91 to 94 percent of the control-correct content predictions under
the early band and 86 to 92 percent under the late band, in both corpora, and it
turns 0.2 to 1.5 percent of the control-wrong ones correct. The early band
removes more than the late band in every corpus and model, so each corpus's
early-minus-late damage gap grows with $p_C$, and $p_\Rdom$ exceeds $p_\Ndom$ for
every type and model in \tabref{tab:app_h1_control_split}. The split computes CAI
separately among control-correct and control-wrong targets and combines the two
cells with the Wikitext shares, $p_\Ndom\cai_{\rm correct}+(1-p_\Ndom)\cai_{\rm
wrong}$. This keeps each stratum's CAI and gives both corpora the reference
composition. The strata come from the unablated pass, which neither band
perturbs. Because the strata use the model's own predictions, the split applies to a cell
that has already been measured.

For content words every reweighted interval includes zero, and the paired
difference from the content cell excludes zero in all five instances. Llama is
the one instance whose control-correct content cell has an interval excluding
zero. Among the other three types the reweighted interval lies wholly above zero
for Llama and Gemma numbers and for Llama function words, three of 15 cells, compared with eight among the unsplit type cells, and it lies below zero for Qwen2.5 punctuation,
the one instance whose control-wrong punctuation cell is negative.

\begin{table}[H]
\centering
\small
\caption{The four non-whitespace types of the H=1 mild-resample series, split by
whether the control pass predicts the target at top-1. $p_\Rdom/p_\Ndom$ is the
control-correct share of the type's GSM8K and Wikitext targets, and CAI is the
type cell of \tabref{tab:app_residual_support} and \tabref{tab:app_h1_types},
which give its interval. Correct and wrong give CAI within each stratum. Reweighted is
$p_\Ndom\cai_{\rm correct}+(1-p_\Ndom)\cai_{\rm wrong}$, and Diff.\ is CAI
minus reweighted, paired within each draw. Bold marks a correct, wrong,
reweighted, or Diff.\ value whose 95\% interval excludes zero. The other intervals use 2,000 sequence-bootstrap draws. A value that rounds to zero is printed to its first significant digit.}
\label{tab:app_h1_control_split}
\setlength{\tabcolsep}{3pt}
\begin{tabular}{@{}lrrrrrcr@{}}
\toprule
Model & $p_\Rdom/p_\Ndom$ & CAI & Correct & Wrong & Reweighted & 95\% CI & Diff. \\
\midrule
\multicolumn{8}{@{}l}{\emph{content word}} \\
Qwen2.5-7B & 0.88/0.39 & $+0.0103$ & $-0.0037$ & $-0.0021$ & $-0.0028$ & $[-0.0073, 0.0018]$ & {\boldmath$+0.0131$} \\
Llama-3.2-3B & 0.62/0.37 & $+0.0142$ & {\boldmath$+0.0106$} & $-0.0014$ & $+0.0031$ & $[-0.0008, 0.0069]$ & {\boldmath$+0.0111$} \\
gemma-2-2b & 0.61/0.33 & $+0.0086$ & $-0.0021$ & $+0.0010$ & $-0.00003$ & $[-0.0036, 0.0034]$ & {\boldmath$+0.0086$} \\
OLMo-2-7B & 0.69/0.43 & $+0.0148$ & $+0.0019$ & $+0.0002$ & $+0.0009$ & $[-0.0034, 0.0055]$ & {\boldmath$+0.0139$} \\
Qwen3-1.7B & 0.62/0.31 & $+0.0125$ & $+0.0008$ & $-0.0001$ & $+0.0002$ & $[-0.0029, 0.0034]$ & {\boldmath$+0.0123$} \\
\midrule
\multicolumn{8}{@{}l}{\emph{number}} \\
Qwen2.5-7B & 0.92/0.61 & $+0.0038$ & $-0.0012$ & $-0.0222$ & $-0.0094$ & $[-0.0223, 0.0027]$ & {\boldmath$+0.0132$} \\
Llama-3.2-3B & 0.75/0.41 & $+0.0629$ & {\boldmath$+0.0562$} & $+0.0101$ & {\boldmath$+0.0292$} & $[0.0118, 0.0486]$ & {\boldmath$+0.0337$} \\
gemma-2-2b & 0.83/0.58 & $+0.0445$ & {\boldmath$+0.0392$} & $+0.0102$ & {\boldmath$+0.0271$} & $[0.0187, 0.0353]$ & {\boldmath$+0.0174$} \\
OLMo-2-7B & 0.79/0.45 & $+0.0462$ & {\boldmath$+0.0356$} & $-0.0148$ & $+0.0080$ & $[-0.0095, 0.0236]$ & {\boldmath$+0.0382$} \\
Qwen3-1.7B & 0.84/0.56 & $+0.0293$ & {\boldmath$+0.0298$} & {\boldmath$-0.0136$} & $+0.0109$ & $[-0.0004, 0.0217]$ & {\boldmath$+0.0184$} \\
\midrule
\multicolumn{8}{@{}l}{\emph{punctuation}} \\
Qwen2.5-7B & 0.81/0.41 & $+0.0007$ & $-0.0104$ & {\boldmath$-0.0801$} & {\boldmath$-0.0513$} & $[-0.0705, -0.0329]$ & {\boldmath$+0.0520$} \\
Llama-3.2-3B & 0.69/0.51 & $+0.0073$ & $-0.0119$ & $+0.0065$ & $-0.0028$ & $[-0.0114, 0.0064]$ & {\boldmath$+0.0101$} \\
gemma-2-2b & 0.75/0.38 & $+0.0234$ & $+0.0147$ & $+0.0041$ & $+0.0081$ & $[-0.0012, 0.0170]$ & {\boldmath$+0.0154$} \\
OLMo-2-7B & 0.68/0.51 & $+0.0183$ & $+0.0128$ & $+0.0042$ & $+0.0086$ & $[-0.0020, 0.0193]$ & {\boldmath$+0.0097$} \\
Qwen3-1.7B & 0.72/0.43 & $+0.0253$ & $+0.0082$ & $+0.0044$ & $+0.0060$ & $[-0.0032, 0.0144]$ & {\boldmath$+0.0193$} \\
\midrule
\multicolumn{8}{@{}l}{\emph{function word}} \\
Qwen2.5-7B & 0.83/0.56 & $+0.0039$ & $-0.0049$ & $-0.0028$ & $-0.0040$ & $[-0.0099, 0.0022]$ & {\boldmath$+0.0079$} \\
Llama-3.2-3B & 0.57/0.53 & $+0.0101$ & {\boldmath$+0.0153$} & $-0.0007$ & {\boldmath$+0.0077$} & $[0.0028, 0.0130]$ & {\boldmath$+0.0023$} \\
gemma-2-2b & 0.53/0.51 & $-0.0004$ & {\boldmath$-0.0082$} & {\boldmath$+0.0070$} & $-0.0007$ & $[-0.0048, 0.0034]$ & $+0.0002$ \\
OLMo-2-7B & 0.59/0.52 & $+0.0025$ & $+0.0060$ & {\boldmath$-0.0122$} & $-0.0028$ & $[-0.0096, 0.0038]$ & {\boldmath$+0.0053$} \\
Qwen3-1.7B & 0.57/0.54 & $-0.0030$ & $-0.0074$ & $+0.0019$ & $-0.0031$ & $[-0.0079, 0.0016]$ & $+0.0001$ \\
\bottomrule
\end{tabular}
\end{table}

Reweighting by strata that depend on the input alone leaves a smaller positive
content cell (\tabref{tab:app_h1_strata}). Repeated tokens, which the model can copy from
earlier in the sequence, make up 0.52 to 0.54 of the GSM8K content targets and
0.18 of the Wikitext ones. With the four repeated-by-leading-space strata weighted by the Wikitext shares, the estimate falls, and its interval stays above zero in
four of five instances, all but Gemma. The high-frequency tertile has an interval
above zero in all five instances, and no low-frequency interval lies above
zero.

\begin{table}[H]
\centering
\small
\caption{Content-word CAI of the H=1 mild-resample series within strata that
depend on the input alone, in bold when the 95\% interval of 2,000
sequence-bootstrap draws excludes zero. A target is repeated when its token
occurs earlier in the same sequence, and a leading space marks a word start.
Frequency tertiles rank each token by its count among the content targets of
both corpora. The last two rows combine the four repeated-by-leading-space
strata with the GSM8K and with the Wikitext stratum shares.}
\label{tab:app_h1_strata}
\setlength{\tabcolsep}{4pt}
\begin{tabular}{@{}lrrrrr@{}}
\toprule
Content stratum & Qwen2.5-7B & Llama-3.2-3B & gemma-2-2b & OLMo-2-7B & Qwen3-1.7B \\
\midrule
all content targets & {\boldmath$+0.0103$} & {\boldmath$+0.0142$} & {\boldmath$+0.0086$} & {\boldmath$+0.0148$} & {\boldmath$+0.0125$} \\
\midrule
repeated in the sequence & {\boldmath$+0.0099$} & {\boldmath$+0.0186$} & {\boldmath$+0.0106$} & $+0.0060$ & {\boldmath$+0.0113$} \\
first occurrence & {\boldmath$+0.0051$} & $+0.0028$ & $+0.0012$ & {\boldmath$+0.0076$} & $+0.0014$ \\
leading space & {\boldmath$+0.0089$} & {\boldmath$+0.0158$} & {\boldmath$+0.0090$} & {\boldmath$+0.0168$} & {\boldmath$+0.0124$} \\
no leading space & {\boldmath$+0.0214$} & $+0.0003$ & $-0.0021$ & $-0.0021$ & {\boldmath$+0.0191$} \\
\midrule
repeated, leading space & $+0.0052$ & {\boldmath$+0.0196$} & {\boldmath$+0.0103$} & $+0.0081$ & {\boldmath$+0.0116$} \\
repeated, no leading space & {\boldmath$+0.0352$} & $+0.0089$ & $-0.0032$ & $+0.0078$ & {\boldmath$+0.0378$} \\
first occurrence, leading space & {\boldmath$+0.0053$} & $+0.0038$ & $+0.0014$ & {\boldmath$+0.0094$} & $+0.0025$ \\
first occurrence, no leading space & $+0.0039$ & $-0.0070$ & $-0.0027$ & $-0.0084$ & $-0.0051$ \\
\midrule
occurs in both corpora & {\boldmath$+0.0099$} & {\boldmath$+0.0154$} & {\boldmath$+0.0089$} & {\boldmath$+0.0139$} & {\boldmath$+0.0132$} \\
occurs in one corpus & {\boldmath$+0.0102$} & {\boldmath$+0.0177$} & {\boldmath$+0.0075$} & {\boldmath$+0.0208$} & {\boldmath$+0.0108$} \\
\midrule
low-frequency tertile & $-0.0002$ & $-0.0012$ & {\boldmath$-0.0044$} & $-0.0037$ & $+0.0030$ \\
middle-frequency tertile & $+0.0028$ & {\boldmath$+0.0072$} & $+0.0032$ & $+0.0048$ & {\boldmath$+0.0072$} \\
high-frequency tertile & {\boldmath$+0.0150$} & {\boldmath$+0.0223$} & {\boldmath$+0.0127$} & {\boldmath$+0.0278$} & {\boldmath$+0.0177$} \\
\midrule
four strata, GSM8K shares & {\boldmath$+0.0070$} & {\boldmath$+0.0109$} & $+0.0053$ & {\boldmath$+0.0078$} & {\boldmath$+0.0085$} \\
four strata, Wikitext shares & {\boldmath$+0.0064$} & {\boldmath$+0.0047$} & $+0.0022$ & {\boldmath$+0.0068$} & {\boldmath$+0.0042$} \\
\bottomrule
\end{tabular}
\end{table}

\subsection{Cross-domain activation patching}

\tabref{tab:app_beyond_cai_b4} reports target-type bootstrap intervals for
activation patching in the three primary models, and \figref{fig:scope}c shows
the same intervention under top-1. Here, activation patching and a
log-probability readout replace ablation and top-1 CAI, respectively. The
separation between same-domain and cross-domain patching further distinguishes
the content/non-whitespace result from the H=3 mean-blend whitespace cell.

\clearpage
\section{Controls and robustness details}
\label{app:controls}

\subsection{Control summary}

\begin{table}[H]
\centering
\small
\caption{Controls supporting the audit interpretation. Numerators report how
many tested primary models satisfy the named criterion.}
\label{tab:app_controls}
\begin{tabular}{>{\raggedright\arraybackslash}p{0.21\linewidth}
                >{\raggedright\arraybackslash}p{0.21\linewidth}
                >{\raggedright\arraybackslash}p{0.44\linewidth}}
\toprule
Control & Result & Interpretation \\
\midrule
Easy/hard GSM8K split at H=3 top-1 & 3/3 artifact pattern in each split & Both splits show the artifact pattern, with no monotonic difficulty gradient. \\
Shuffled-rationale H=3 top-1 & 0/3 positive content pattern & With the rationale shuffled and its token distribution preserved, no model shows a positive H=3 content-token pattern, the same sign as the canonical H=3 content cells of \tabref{tab:canonical}. \\
Shuffled-rationale H=3 logprob & 0/3 positive content pattern & The null also appears under the log-probability readout. \\

Norm-equalized H=3 top-1 & 3/3 pooled positive & After equalizing perturbation norms across domains, pooled CAI is +0.027/+0.040/+0.040 in Qwen/Llama/Gemma. \\
\bottomrule
\end{tabular}
\end{table}

\subsection{Base-checkpoint robustness}

\begin{table}[H]
\centering
\small
\caption{Pooled H=3 top-1 CAI of base checkpoints on GSM8K/Wikitext under
mean-blend at $\alpha=0.2$, with 95\% bootstrap intervals and $n$
length-matched sequences per corpus. The rows name base releases, which carry
no \texttt{-Instruct} or \texttt{-it} suffix.}
\label{tab:app_base}
\begin{tabular}{lrrr}
\toprule
Base checkpoint & $n$ & pooled H=3 CAI & 95\% CI \\
\midrule
OLMo-2-1124-7B & 298 & +0.0342 & [0.0298, 0.0388] \\
Qwen2.5-7B & 290 & +0.0237 & [0.0192, 0.0279] \\
gemma-2-2b & 285 & +0.0107 & [0.0058, 0.0152] \\
\bottomrule
\end{tabular}
\end{table}

The positive pooled H=3 estimate appears before instruction tuning in all three
base checkpoints. These base-checkpoint estimates pool over target types, and the target-type
decomposition uses the instruction-tuned checkpoints.

\subsection{Perturbation strength}
\label{app:alpha}

Every estimate outside this section fixes the perturbation strength at
$\alpha=0.2$. \tabref{tab:app_alpha} reports a mild-resample strength sweep:
pooled top-1 CAI under mild-resample on GSM8K/Wikitext at seven values of
$\alpha$ from $0.05$ to $0.40$, for the three primary models. With a positive cell defined as a strictly positive 95\% interval, H=1 gives four to six positive cells
of the seven strengths across the three models, and H=3 gives at most two. Qwen's H=1 interval stops excluding zero above $\alpha=0.2$, and at
$\alpha=0.1$ the Llama and Gemma H=1 estimates are strictly negative, so the
H=1 estimate varies with strength in a pattern that differs across models.

\begin{table}[H]
\centering
\small
\caption{Pooled top-1 CAI under mild-resample on GSM8K/Wikitext across
perturbation strengths, with 200 sequences per corpus and 1,000 sequence-level
bootstrap resamples. Values carry five decimals because several H=3 interval
endpoints fall below $10^{-4}$. The $\alpha=0.20$ rows are the setting used
elsewhere in the paper, measured on this 200-sequence subsample.}
\label{tab:app_alpha}
\begin{tabular}{lrrrr}
\toprule
$\alpha$ & H=1 CAI & H=1 95\% CI & H=3 CAI & H=3 95\% CI \\
\midrule
\multicolumn{5}{l}{\textit{Qwen2.5-7B}} \\
\quad 0.05 & +0.08698 & [0.07763, 0.09647] & \textminus0.00148 & [\textminus0.00334, 0.00021] \\
\quad 0.10 & +0.10093 & [0.08539, 0.11567] & \textminus0.00649 & [\textminus0.00911, \textminus0.00373] \\
\quad 0.15 & +0.00970 & [0.00411, 0.01469] & \textminus0.00132 & [\textminus0.00260, \textminus0.00002] \\
\quad 0.20 & +0.00531 & [0.00147, 0.00933] & \textminus0.00094 & [\textminus0.00201, 0.00014] \\
\quad 0.25 & +0.00296 & [\textminus0.00048, 0.00639] & \textminus0.00045 & [\textminus0.00136, 0.00044] \\
\quad 0.30 & +0.00130 & [\textminus0.00207, 0.00483] & \textminus0.00081 & [\textminus0.00174, 0.00002] \\
\quad 0.40 & \textminus0.00137 & [\textminus0.00439, 0.00180] & \textminus0.00085 & [\textminus0.00173, \textminus0.00002] \\
\midrule
\multicolumn{5}{l}{\textit{Llama-3.2-3B}} \\
\quad 0.05 & +0.01308 & [0.00667, 0.01970] & +0.00314 & [0.00123, 0.00493] \\
\quad 0.10 & \textminus0.03102 & [\textminus0.04428, \textminus0.01797] & \textminus0.00547 & [\textminus0.00851, \textminus0.00218] \\
\quad 0.15 & +0.02514 & [0.02004, 0.03071] & +0.00172 & [0.00015, 0.00305] \\
\quad 0.20 & +0.01780 & [0.01239, 0.02345] & \textminus0.00055 & [\textminus0.00140, 0.00029] \\
\quad 0.25 & +0.01413 & [0.00901, 0.01946] & \textminus0.00020 & [\textminus0.00089, 0.00050] \\
\quad 0.30 & +0.01256 & [0.00786, 0.01758] & \textminus0.00023 & [\textminus0.00089, 0.00043] \\
\quad 0.40 & +0.01101 & [0.00607, 0.01588] & \textminus0.00020 & [\textminus0.00086, 0.00044] \\
\midrule
\multicolumn{5}{l}{\textit{gemma-2-2b}} \\
\quad 0.05 & +0.02397 & [0.01869, 0.02954] & +0.00176 & [0.00004, 0.00335] \\
\quad 0.10 & \textminus0.05301 & [\textminus0.06518, \textminus0.04073] & \textminus0.00335 & [\textminus0.00618, \textminus0.00029] \\
\quad 0.15 & +0.02132 & [0.01591, 0.02644] & \textminus0.00046 & [\textminus0.00172, 0.00069] \\
\quad 0.20 & +0.01559 & [0.01062, 0.02078] & \textminus0.00053 & [\textminus0.00149, 0.00046] \\
\quad 0.25 & +0.01176 & [0.00762, 0.01596] & \textminus0.00038 & [\textminus0.00126, 0.00045] \\
\quad 0.30 & +0.00892 & [0.00495, 0.01292] & \textminus0.00041 & [\textminus0.00122, 0.00037] \\
\quad 0.40 & +0.00495 & [0.00173, 0.00821] & \textminus0.00058 & [\textminus0.00127, 0.00007] \\
\bottomrule
\end{tabular}
\end{table}

\tabref{tab:app_alpha_meanblend} measures the strength dependence of the H=3
mean-blend decomposition itself. It applies the early- and late-band passes to the canonical 300-sequence sample
at five strengths, and at $\alpha=0.2$ it reproduces the pooled value, the five conditional cells, and both reweighted
summaries of \tabref{tab:canonical}. The pooled interval lies above zero in 13 of the
15 model--strength rows, all but Qwen and Gemma at $\alpha=0.05$. The whitespace
interval lies above zero at every strength from $0.10$ in every model, and at
$0.05$ for Llama. The directly measured non-whitespace interval lies above zero
in none of the 15 rows and below zero in 11 of the 13 pooled-positive rows. The one positive whitespace-excluded summary is Gemma's at $\alpha=0.10$, where
the direct non-whitespace interval, [$-0.0016$, $0.0016$], contains zero.

\begin{table}[H]
\centering
\small
\caption{H=3 mean-blend top-1 CAI on the canonical GSM8K/Wikitext sample across
perturbation strengths, with 300 sequences per corpus and 2,000 sequence-level
bootstrap resamples. The columns are the pooled value, the whitespace cell, the
whitespace-excluded summary with combined-count weights, and the non-whitespace
CAI measured directly. At $\alpha=0.20$ the estimates equal those of
\tabref{tab:canonical}, and the intervals come from this sweep's resamples. A value that rounds to zero is printed to its first significant
digit.}
\label{tab:app_alpha_meanblend}
\setlength{\tabcolsep}{2.5pt}
\resizebox{\linewidth}{!}{%
\begin{tabular}{lrrrrrrrr}
\toprule
$\alpha$ & pooled & 95\% CI & ws & 95\% CI & ws-excl. & 95\% CI & nonws direct & 95\% CI \\
\midrule
\multicolumn{9}{l}{\textit{Qwen2.5-7B}} \\
\quad 0.05 & \textminus0.0008 & [\textminus0.0022, 0.0005] & \textminus0.0157 & [\textminus0.0390, 0.0025] & \textminus0.0017 & [\textminus0.0032, \textminus0.0004] & \textminus0.0015 & [\textminus0.0028, \textminus0.0003] \\
\quad 0.10 & +0.0025 & [0.0005, 0.0043] & +0.0514 & [0.0240, 0.0753] & \textminus0.0050 & [\textminus0.0067, \textminus0.0034] & \textminus0.0047 & [\textminus0.0063, \textminus0.0031] \\
\quad 0.20 & +0.0253 & [0.0213, 0.0292] & +0.5181 & [0.4770, 0.5562] & \textminus0.0247 & [\textminus0.0272, \textminus0.0220] & \textminus0.0253 & [\textminus0.0280, \textminus0.0227] \\
\quad 0.30 & +0.0262 & [0.0213, 0.0310] & +0.8573 & [0.8323, 0.8797] & \textminus0.0529 & [\textminus0.0557, \textminus0.0499] & \textminus0.0511 & [\textminus0.0541, \textminus0.0481] \\
\quad 0.40 & +0.0284 & [0.0239, 0.0328] & +0.8999 & [0.8791, 0.9192] & \textminus0.0543 & [\textminus0.0574, \textminus0.0509] & \textminus0.0516 & [\textminus0.0544, \textminus0.0489] \\
\midrule
\multicolumn{9}{l}{\textit{Llama-3.2-3B}} \\
\quad 0.05 & +0.0020 & [0.0007, 0.0032] & +0.0229 & [0.0094, 0.0351] & \textminus0.0005 & [\textminus0.0016, 0.0006] & \textminus0.0009 & [\textminus0.0019, 0.0002] \\
\quad 0.10 & +0.0212 & [0.0183, 0.0241] & +0.2698 & [0.2401, 0.3014] & \textminus0.0062 & [\textminus0.0082, \textminus0.0042] & \textminus0.0078 & [\textminus0.0097, \textminus0.0058] \\
\quad 0.20 & +0.0428 & [0.0383, 0.0471] & +0.8215 & [0.7883, 0.8521] & \textminus0.0330 & [\textminus0.0355, \textminus0.0303] & \textminus0.0329 & [\textminus0.0355, \textminus0.0303] \\
\quad 0.30 & +0.0309 & [0.0264, 0.0354] & +0.6593 & [0.6219, 0.6944] & \textminus0.0396 & [\textminus0.0429, \textminus0.0360] & \textminus0.0386 & [\textminus0.0417, \textminus0.0350] \\
\quad 0.40 & +0.0344 & [0.0297, 0.0396] & +0.7287 & [0.6930, 0.7657] & \textminus0.0528 & [\textminus0.0563, \textminus0.0491] & \textminus0.0503 & [\textminus0.0535, \textminus0.0470] \\
\midrule
\multicolumn{9}{l}{\textit{gemma-2-2b}} \\
\quad 0.05 & \textminus0.0006 & [\textminus0.0016, 0.0005] & \textminus0.0073 & [\textminus0.0245, 0.0078] & \textminus0.0009 & [\textminus0.0019, 0.0000] & \textminus0.0014 & [\textminus0.0024, \textminus0.0004] \\
\quad 0.10 & +0.0087 & [0.0068, 0.0108] & +0.0751 & [0.0504, 0.0968] & +0.0018 & [0.0001, 0.0034] & \textminus0.00003 & [\textminus0.0016, 0.0016] \\
\quad 0.20 & +0.0421 & [0.0380, 0.0462] & +0.7155 & [0.6875, 0.7402] & \textminus0.0131 & [\textminus0.0157, \textminus0.0104] & \textminus0.0155 & [\textminus0.0186, \textminus0.0125] \\
\quad 0.30 & +0.0236 & [0.0189, 0.0285] & +0.8763 & [0.8628, 0.8895] & \textminus0.0512 & [\textminus0.0539, \textminus0.0485] & \textminus0.0518 & [\textminus0.0550, \textminus0.0485] \\
\quad 0.40 & +0.0329 & [0.0286, 0.0374] & +0.9155 & [0.9040, 0.9262] & \textminus0.0497 & [\textminus0.0523, \textminus0.0472] & \textminus0.0456 & [\textminus0.0481, \textminus0.0433] \\
\bottomrule
\end{tabular}}
\end{table}

\clearpage
\section{Recommended reporting checklist for future studies}
\label{app:future_checklist}

For any naturalistic domain-pair depth-attribution study, we recommend
reporting the following before assigning a broad mechanistic interpretation
to an aggregate effect:

\begin{enumerate}[leftmargin=*]
\item Log the exact target token ID and decoded form beside every score or
correctness bit, together with sequence and source-position identifiers.
\item Report the domain-pair target-token-type distributions
$\pi_{\Rdom}$ and $\pi_{\Ndom}$ at every evaluated horizon.
\item Name the checkpoint, intervention site and operator, layer bands,
perturbation strength, horizon, and readout for each claim.
\item Compute a low-cost prospective risk proxy, such as characters-per-token
asymmetry, before inspecting intervention outcomes.
\item Report conditional effects $\cai^o_{k,\tau}$ for whitespace, content,
number, punctuation, and function-word targets, including cell support and
sequence-level uncertainty.
\item Report a whitespace-excluded or otherwise task-appropriate mixture
estimate and a reference-reweighted estimate.
\item When target-token asymmetry is non-trivial, compare the suspected
artifact cell with a task-appropriate calibration across operators or readouts.
\item Report every positive cell that remains, and split it by the unablated
model's top-1 correctness on its targets, since a double difference of damage
grows with control accuracy when the corpora differ in it.
\item Add a note on domain semantics for code, markup, tables, or other formats
where whitespace or punctuation is structural.
\item Report evaluated sequence counts and numbers of bootstrap resamples.
\end{enumerate}

This checklist provides the minimum information needed to interpret an
aggregate depth claim over naturalistic domain pairs. An effect that
survives the audit supports a positive finding for the audited target types, and an effect that
collapses under the audit identifies a mixture-driven aggregate.

\clearpage
\section{Reproducibility details}
\label{app:repro}

The main text and appendix provide the measurement protocol and report the
numerical results for the canonical decomposition, held-out and reasoning-pair
extensions, DoLa-motivated corpus audit, classifier-policy analysis, mechanism
and margin analyses, and position control. The paper reports summary statistics,
uncertainty intervals, and sample counts for these analyses. The underlying
research code, machine-readable result files, and figure-generation scripts are
not part of this arXiv submission.

\noindent\emph{Function-word inventory for Appendix A.2.} The exact 106
function words, in lower case, are:
{\footnotesize\texttt{the, a, an, is, are, was, were, be, been, being, have, has, had, do, does, did, will, would, could, should, may, might, can, shall, must, to, of, in, for, on, with, at, by, from, as, into, through, during, before, after, above, below, between, out, and, but, or, nor, not, so, yet, both, either, neither, each, every, all, any, few, more, most, other, some, such, no, only, same, than, too, very, just, that, this, these, those, it, its, he, she, they, them, their, we, us, our, you, your, i, me, my, if, then, else, when, while, where, who, which, what, how, there, here, also, about, up, over}.}

Each uncertainty statement identifies the sequence as the resampling unit. The MATH/DROP/ARC audits use 2,000 resamples.
The five-instance interaction, DoLa, classifier-policy, and position audits use
10,000. The mixture-summary intervals of \tabref{tab:canonical} and
\tabref{tab:app_expansion_intervals}, DoLa rows included, use 2,000. Table~\ref{tab:coverage},
Table~\ref{tab:app_heldout_outcomes},
Table~\ref{tab:app_reasoning_extension}, and Table~\ref{tab:app_dola} give every matched sample size.

\clearpage
\section{Broader impact and data considerations}
\label{app:impact}

The audit lets an intervention-based analysis of a language model check whether
an aggregate depth effect depends on the target mixture. Applying it
to an aggregate calls for both parts of the paired reporting rule: the mixture
correction on the pooled estimate, and a report of the target-type cells that
remain positive, split by the unablated model's correctness on their targets. The
two together keep a corrected aggregate and its remaining positive cells in view
at once. This arXiv submission contains no research scripts,
machine-readable result files, or model weights.

\section{Numerical summary for the position figure}
\label{app:figure_tables}

\figref{fig:position} presents estimates that are also useful in numerical form.
The table below reports the plotted values at the precision used in the text. The
operator $\times$ horizon grid of \figref{fig:scope} is tabulated next to that
figure, in \tabref{tab:app_operator_horizon_grid}.

\begin{table}[H]
\centering
\small
\caption{Whitespace H=3 CAI by relative-position quartile and after position
matching, as shown in \figref{fig:mechanism}c and \figref{fig:position}c. Values are point
estimates, and all 15 intervals exclude zero. The full
intervals, event counts, and answer-distance bins are in
Appendix~\ref{app:position_control}.}
\label{tab:position_main}
\begin{tabular}{lrrrrr}
\toprule
Model & Q1 & Q2 & Q3 & Q4 & position-matched \\
\midrule
Qwen2.5-7B & $+0.422$ & $+0.492$ & $+0.460$ & $+0.431$ & $+0.447$ \\
Llama-3.2-3B & $+0.612$ & $+0.919$ & $+0.842$ & $+0.810$ & $+0.791$ \\
gemma-2-2b-it & $+0.351$ & $+0.663$ & $+0.786$ & $+0.762$ & $+0.626$ \\
\bottomrule
\end{tabular}
\end{table}

\end{document}